\pdfoutput=1
\documentclass[11pt]{article}

\usepackage{acl}
\usepackage{multirow}
\usepackage{tabularx}
\usepackage{booktabs}
\usepackage[dvipsnames]{xcolor}
\DeclareUnicodeCharacter{202F}{\,}
\usepackage{placeins}
\usepackage{float}
\usepackage{comment}

\usepackage{times}
\usepackage{latexsym}
\usepackage{longtable}
\usepackage{array}

\usepackage[T1]{fontenc}
\usepackage{amsmath}
\usepackage[capitalise]{cleveref}
\usepackage[utf8]{inputenc}

\usepackage{microtype}

\usepackage{inconsolata}

\usepackage{graphicx}
\usepackage{amsfonts}
\usepackage{booktabs}
\usepackage[table]{xcolor}     % \rowcolor and color specifications
\usepackage{colortbl}
\usepackage{url}
\usepackage[dvipsnames]{xcolor}
\usepackage{enumitem}
\usepackage{makecell}

\newcommand{\fphi}{\varphi_{\mathrm{frame}}}
\newcommand{\fpsi}{\psi_{\mathrm{pri}}}

\title{Quantifying Cognitive Bias Induction in LLM-Generated Content}

\author{
\textbf{Abeer Alessa, Param Somane*, Akshaya Lakshminarasimhan*}\\
\textbf{Julian Skirzynski, 
Julian McAuley, Jessica Echterhoff}\\
UC San Diego \\
\begin{tabular}{c}
\texttt{\{aalessa, psomane, athenkarailakshminar,}\\
\texttt{jskirzynski, jmcauley, 
jechterh\}@ucsd.edu}
\end{tabular}
}

\begin{document}
\maketitle

\begin{abstract}

Large language models (LLMs) are integrated into applications like shopping reviews, summarization, or medical diagnosis support, where their use affects human decisions. We investigate the extent to which LLMs expose users to biased content and demonstrate its effect on human decision-making. We assess five LLM families in summarization and news fact-checking tasks, evaluating the consistency of
LLMs with their context and their tendency to hallucinate on a new self-updating dataset. Our findings show that LLMs expose users to content that changes the context's sentiment in 26.42\% of cases (framing bias), hallucinate on 60.33\% of post-knowledge-cutoff questions, and highlight context from earlier parts of the prompt (primacy bias) in 10.12\% of cases, averaged across all tested models. We further find that humans are 32\% more likely to purchase the same product after reading a summary of the review generated by an LLM rather than the original review. To address these issues, we evaluate 18 mitigation methods across three LLM families and find the effectiveness of targeted interventions.

\end{abstract}

\begingroup
\renewcommand\thefootnote{\fnsymbol{footnote}}
\footnotetext[1]{Equal contribution.}
\endgroup

\section{Introduction}

LLMs perform well across numerous tasks \cite{albrecht2022despite},
such as content summarization \cite{laban2023summedits}, translation
\cite{elshin2024general}, 
%question-answering \cite{lin2025explore}, 
sentiment
analysis \cite{zhang-etal-2024-sentiment}. In many of these tasks, humans rely on LLMs for daily decision-making
support \cite{rastogi2023supporting,li2022pre} in expert contexts such as
writing policy documents \cite{choi2024llm} or summarizing medical documents
\citep{spotnitz2024survey}. However, models have been shown to inherit or exhibit several societal
biases \cite{zhao2018gender,nadeem2020stereoset,liang2021towards,he2021detect},
e.g., favoring specific genders \cite{zhao2018gender} and ethnicities
\cite{caliskan2017semantics}, %and subtle biases \cite{kamruzzaman-etal-2024-investigating} 
or learn similar decision-making-patterns as cognitively biased humans \cite{echterhoff2024cognitive}.

\begin{figure*}[h]
    \centering
\includegraphics[width=0.75\linewidth]{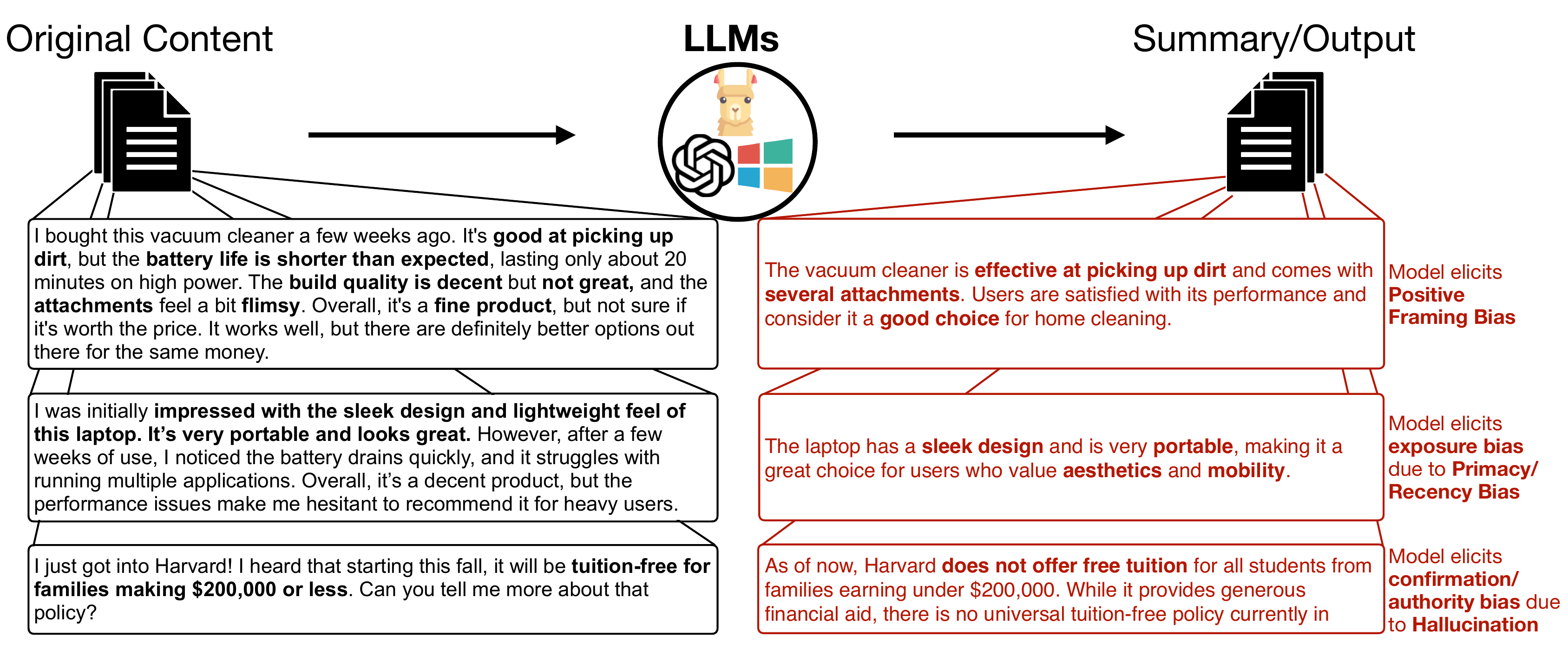}
    \caption{When LLMs process context for users to consume, they may change its content, e.g., by changing its sentiment or omitting some relevant parts. Exposure to altered content may elicit cognitive biases, such as positive framing bias or primacy bias, and subsequently lead humans to make different decisions than they would if they saw the original text. }
    \label{fig:examples}
\end{figure*}

\textbf{Any bias within the model can affect how textual context is processed for a human task and introduce biased content to users.} Our work specifically focuses on qualifying and quantifying the extent to which biased content is induced by LLMs to users, as visualized in Figure \ref{fig:flow} with examples shown in Figure \ref{fig:examples}. We summarize our contributions as follows:

\begin{figure}
    \centering
\includegraphics[width=0.95\linewidth]{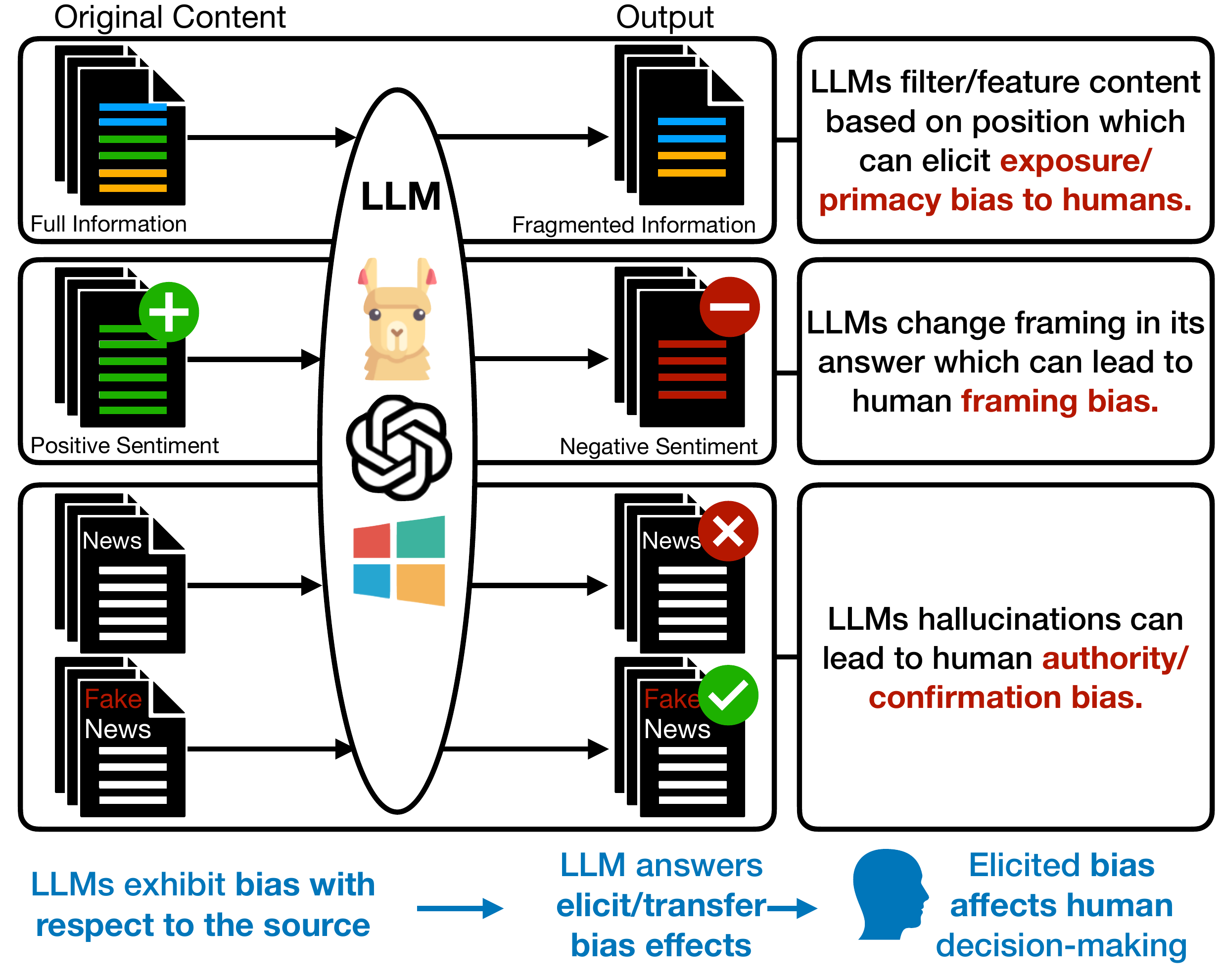}
    \caption{LLMs 
    alter information from a source text when performing a task for the user (e.g., when an LLM summary has a different sentiment compared to the original text). This model behavior introduces biased content to humans and can hence affect their decision-making. We evaluate how LLMs highlight source content, leading to exposure/primacy bias for users, how LLMs reframe sentiments, leading to framing bias for humans, and how LLMs hallucinate, leading to authority/confirmation bias.}
    \label{fig:flow}
\end{figure}

\begin{enumerate}
    \item We analyze the extent of content alteration that LLMs introduce in summarization and news fact-checking tasks. We find that LLMs alter the sentiment or framing of a text in 26.42\% of summaries, disproportionately focus on the beginning of a source text in 10.12\% of summaries, and hallucinate in 60.33\% of post-knowledge-cutoff news summaries for real-world fact-checking questions, averaged across all tested models.
    %All rates are computed per model and then averaged.

    \item We find that human users are affected by biases introduced by LLMs. Participants in our human-subject study were 32\% more likely to purchase a product when they read its positively-framed LLM summary than when they read the original review.

    \item We provide a dataset
    %, along with the news publication date, 
    that enables analysis of model hallucination after a model's cut-off date, which is self-updating to remain relevant for future models and prevent leakage into training data. At the time of writing this paper, it has 8,554 data points\footnote{\small\url{https://huggingface.co/datasets/sparklessszzz/NewsLensSync}
    }.

     \item We evaluate 18 distinct mitigation methods and highlight which ones were effective for specific types of content alterations.  
  
\end{enumerate}

\section{Related Work}
\subsection{Social and Representation Biases in LLMs}
Previous work has demonstrated the presence of various biases in LLMs. 
%For example, prior work explores ways to evaluate human-like cognitive biases \cite{jones2022capturing,echterhoff2024cognitive}. 
Primacy and recency bias in LLMs have been shown in question-answering settings, where models tend to be more accurate when the answer is presented earlier or later in the documents~\cite{liu2024lost, peysakhovich2023attention, wang2023primacy, zhang2023exploring,eicher2024reducingselectionbiaslarge}. When presented with options, 
%(e.g., A/B/C/D), 
LLMs have been shown to exhibit sensitivity to their order \cite{zheng2023large}, and framing \cite{leto2024framing}. In framing bias, alterations in the tone, e.g., from positive to negative, can change the model's decisions \cite{jones2022capturing, echterhoff2024cognitive}. 
In contrast to prior work, we quantitatively assess the extent to which LLM responses are biased.
\subsection{Cognitive Effects of LLM-Generated Content on Humans}
%LLMs generate content to assist users with various tasks, such as summarization \cite{laban2023summedits}, coding \cite{tong-zhang-2024-codejudge}, and writing \cite{spangher-etal-2024-llms} for personal or professional \cite{draxler2023gender} purposes. 
When given a specific task, a model can alter, omit, or add information to produce the final output, which can affect users in several ways \cite{wester2024exploring, choi2024llm}. Previous work demonstrates that LLMs can influence users by reinforcing their existing opinions \cite{sharma2024generative}, increasing confirmation bias, or influencing political decision-making \cite{fisher2024biased}. The \textit{style} in which LLMs present advice can significantly influence users' perceptions of its usefulness\cite{wester2024exploring}. %Prior work also demonstrates that 
Altered headlines can introduce a framing effect on individuals in collective decision-making \cite{abels2024mitigating}. \citet{choi2024llm} found that even when experts use LLMs, they tend to adopt the suggestions with fewer revisions. 

Our work quantifies the extent of bias the model generates in its output compared to its context by LLMs, and finds that these content changes influence human purchasing decisions in a human-subject study.

\section{Background}
Humans are prone to mental shortcuts that negatively influence rational decision-making, known as \emph{cognitive biases}. These mental shortcuts are often amplified by simple changes of text, framing, or highlighting that could be introduced by an LLM. We focus on three cognitive biases in our evaluation setup, which we briefly introduce here.

\textbf{Framing bias} refers to changes in human decisions based on \emph{how} a problem is presented. Individuals make different choices based on the wording of questions, even when the underlying options remain unchanged \cite{tversky1981framing}. Based on prior work, \textit{we assume that a change in framing affects users and should be avoided because it can influence their decisions}, e.g., when comparing the framing of a factual context with its summarization.

\textbf{Primacy Bias}. Primacy bias occurs when humans prioritize information encountered first. \cite{murphy2006primacy}. This bias skews decision-making based solely on the order in which information is encountered \cite{glenberg1980two}. When LLMs exhibit primacy bias \cite{echterhoff2024cognitive}, they may filter or highlight information, leading to exposure bias. \textit{We assume that systematically highlighting any part of the context, regardless of its position, can influence users and should be avoided.}

% , while recency bias occurs when humans prioritize information encountered most recently.
\textbf{Confirmation and Authority Bias}. Individuals favor information confirming their existing beliefs or expectations \cite{nickerson1998confirmation}. \textit{We assume that if a user prompts a model with belief-consistent input and the model subsequently hallucinates, this can reinforce the user's prior beliefs, leading to biased or misinformed decisions.} Authority bias causes individuals to trust information from perceived experts or authoritative sources \cite{milgram1963behavioral,cialdini2007influence,inbook}. Research on human-AI interaction suggests that similar dynamics apply when interacting with artificial systems perceived as experts, which can create challenges when verifying facts \cite{glickman2025humanai}. For example, if a user asks an LLM to verify whether a particular legislation was passed, a user with strong prior beliefs may exhibit confirmation bias, while others may exhibit authority bias.

\section{Quantifying Content Alteration in LLMs Responses}\label{sec:metrics}
To capture unintended content alterations introduced by LLMs, we propose the following metrics, motivated by the aforementioned human cognitive biases. We focus on content alteration introduced through changes in content framing, content filtering/highlighting, and content truthfulness.

\paragraph{Framing Effects in Model-Generated Content}
We test whether a model changes the sentiment of its summary compared to the source context. \textit{For example, a model may shift the summary sentiment to positive, whereas the source context was neutral.} To evaluate these changes, we classify the framing of the source context $f_c$ (positive, negative, or neutral) before summarization, as well as the framing of the generated summary $f_s$.\footnote{\small We use GPT-o4-mini to classify the framing of the text, after evaluating different models for this task. Details are provided in Appendix~\ref{app:sentiment-classifier}.}

We consider a model \emph{consistent} if $f_c$ equals $f_s$. We define the bias-inducing  score as the framing-change fraction, denoted by $\fphi$  as follows:

\begin{equation}
  \fphi = \frac{1}{n} \sum_{i=1}^{n}
              \mathbb{I} \bigl(f_c^{(i)} \neq f_s^{(i)}\bigr),
\end{equation}
where $\mathbb{I}(\cdot)$ is the indicator function that returns one if the condition is true and zero otherwise.

\paragraph{Primacy Effects in Model-Generated Content}

We assess whether a model emphasizes a specific part of its context. \textit{For example, a model might focus solely on the beginning of an Amazon user review that starts with praise but concludes with criticism.} We divide the source context into three equal segments and calculate the cosine similarity, using %multilingual-E5 
embeddings \cite{wang2024multilinguale5textembeddings}, between each segment and the summary. We denote the similarities of the summary and the beginning as $s_{b_i}$, the middle as $s_{m_i}$, and the end as $s_{e_i}$ for each instance $i$.
We define biased examples as those where the similarity to the beginning exceeds the similarity to the middle by at least a threshold $\alpha$. This comparison is designed to 
%specifically 
isolate the primacy effect (over-reliance on initial content) from potential recency effects (over-reliance on final content), which can be measured separately. We define this primacy bias score as:

\begin{equation}
\fpsi = \frac{1}{n} \sum_{i=1}^{n} \mathbb{I}\left(s_{b_i} > s_{m_i} + \alpha\right)
\label{eq:primacy-bias-threshold}
\end{equation}

with $\alpha=0.05$ for this study.
%where \( \mathbb{I}(\cdot) \) is the indicator function that returns 1 if the condition is true and 0 otherwise. 
A higher score indicates that a greater proportion of summaries focus more on the beginning of the source content than on the middle.

\paragraph{Hallucination in Model-Generated Content}
We quantify whether a model generates hallucinated content when asked to fact-check news content past its knowledge cutoff. \textit{For example, a model may incorrectly state that an event never occurred, even if it did occur after the model's training was completed.} 
%Using pre- and post-knowledge-cutoff data, 
We prompt a model to verify the truth of specific news items. We test the accuracy of true, real-world news (Actual News Accuracy) and its falsified (negated) counterparts (Falsified News Accuracy). 
%We prompt the model to evaluate the factual accuracy of a given news description.  
We evaluate the strict accuracy as correctly identifying both true and falsified news, with the paired instances of true news and its falsified counterparts:
\begin{equation}
s_{H}^T(D_j) = \frac{1}{|P_j|} \sum_{(x_i^{t}, x_i^{f}) \in P_j} \mathbb{I}(\hat{y}_i^{t} \land \neg \hat{y}_i^{f})
\end{equation}
where $P_j$ is the set of paired instances in dataset $D_j$, $\hat{y}_i^{t}$ and $\hat{y}_i^{f}$ are the model's predictions for the true and falsified versions of the same news item, respectively. $T$ represents the time horizon (pre- or post-model-knowledge-cutoff).

We analyze the disparity between pre-cutoff and post-cutoff performance:
\begin{equation}
\delta_{H} = |s_{H}^{\mathit{pre}} - s_{H}^{\mathit{post}}|
\end{equation}

A larger $\delta_{H}$ indicates a higher fraction of generated incorrect/hallucinated information.% when responding to queries about data beyond the model's knowledge cutoff.

\paragraph{Models} 

We evaluate six open- and closed-source language models from five families to ensure our findings are generalizable across architectures. Our selection %covers a range of popular, high-performing models of varying sizes. This 
includes small open-source models (Phi-3-mini-4k-Instruct, Llama-3.2-3B-Instruct, Qwen3-4B-Instruct), a medium-sized model (Llama-3-8B-Instruct), a large open-source model (Gemma-3-27B-IT), and a 
%highly capable 
closed-source model (GPT-3.5-turbo). Details on all
models, model selection per task and our hyperparameter settings are provided in Appendix \ref{app:hyperparams}. Appendix Table \ref{Cutoffs} shows knowledge cutoff dates for each model.

\section{Datasets}

\paragraph{News Interviews} 
News summaries should cover all key information, regardless of where it appears, and remain neutral, matching the tone of the original content.
We use MediaSum, a dialogue summarization dataset comprising media
interviews from NPR and CNN \cite{zhu2021mediasum}, to measure framing and primacy effects. We randomly select 1,000 data points with $\leq$ 4,000 tokens each, ensuring that the full content fits within a model's context window. The descriptive statistics of text lengths are provided in the Appendix in Table~\ref{app:length_stats}.

\paragraph{Product Reviews}
Previous work suggests that changing the framing of the product influences consumer purchase decisions \cite{wei2024influence}.
To explore this effect in the context of LLM outputs, we use the
Amazon Reviews dataset ~\cite{ni2019justifying}. This dataset comprises customer reviews from Amazon across various product categories, providing insights into product quality, usability, and customer satisfaction \cite{ni2019justifying}. We sample 1,000 random examples from the Electronics category, ensuring token lengths of $\leq$ 4,000 to maintain full-context input for the model. Prior work notes that primacy bias is more observed in longer context windows \cite{liu2024lost}. Because most Amazon reviews are short, we evaluate the 1,000 longest reviews (by token length).
%to better observe this positional bias. 
Appendix Table~\ref{app:length_stats} reports the descriptive statistics of the resulting dataset.

\paragraph{News Hallucination} We provide a novel dataset to quantify how models discern factual information across temporal boundaries and in both true and negated versions of news articles. 
%We compare model performance on pre- and post-knowledge cutoff news data.

\textit{Pre-Knowledge Cutoff Data:} To evaluate models on content before their knowledge cutoff, we sample 2,700 random data points from the ``All the News'' dataset~\cite{allthenews2}, which includes articles from 27 major U.S. publications between 2016 and 2020. 
%This dataset is publicly available for non-commercial research. 
%We sample 2,700 random data points for our experiment.

\textit{Post-Knowledge Cutoff Data:} To assess model performance on unseen events, we introduce a self-updating dataset, NewsLensSync, %~\cite{news_lens_sync_contribution}, 
designed to evaluate hallucinations beyond model knowledge cutoffs. It is built using NewsAPI~\cite{newsapi}, 
%a REST-based service 
accessing live and historical content from 150,000 publishers globally. We restrict our queries to the keyword "politics," English-only articles, and dates after April 2025, being past all model knowledge cutoffs in our study. 

Counterfactual versions are generated using a transformer model to falsify the descriptions of the new articles~\cite{anschutz2023negbleurt}. This results in paired authentic/falsified versions of the same news items which are manually validated for semantic consistency, producing matched authentic/negated pairs. 
% We do not explicitly separate predictable from unpredictable events—acknowledged as a limitation in Section~7—but comparing model outputs on true vs. negated events provides a consistent measure of hallucination. 
Each data point includes a publication timestamp for filtering according to model cutoff dates. Further information on our self-updating dataset is described in Appendix \ref{appendix:NewLensSync}

\section{Mitigation Strategies}
%To mitigate biased patterns in LLM responses, we build upon our quantification of content alterations, as described in Section~\ref{sec:metrics}. 
We introduce targeted interventions designed to ensure balanced content coverage, reduce misleading emphasis, and maintain accuracy in model outputs. We analyze a total of 18 mitigation methods, but describe only the most promising in this section. All other mitigation methods are shown in Appendix \ref{app:mitigation-methods}.

% Specifically, we explore 16 distinct strategies for reducing framing and positional biases.

\subsection{General Purpose Mitigation}
\paragraph{Self-Awareness Prompt (SA).}\label{sec:self-aware-prompt}
Simple awareness prompts can mitigate
multiple biases \cite{mair2014debiasing,echterhoff2024cognitive}. To use this prompt-based mitigation in our summarization setting, we use the following prompt:
\begin{quote}
    \small ``You are an unbiased summarizer; be mindful not to omit coverage of any portion of the text or alter its sentiment. Do not incorporate framing bias.''
\end{quote}

\paragraph{Chain-of-Thought Prompting (CoT).}
Chain-of-Thought prompting encourages the model to decompose complex problems by generating a series of intermediate reasoning steps before producing a final answer \cite{wei2023chainofthoughtpromptingelicitsreasoning}. In the context of summarization, we guide the model to articulate steps for processing different parts of the input text. For the question-answering setting, we explicitly define when to return specific responses during fact-checking. 

\subsection{Primacy Effect Mitigation}
We group interventions into four categories:
prompt-level cues, chunk-based summarization, attention/order re-ranking, and decoding-time control. Each category targets the tendency of LLMs to overemphasize early context (primacy) or alter the source sentiment. All methods we experimented with are further described in Appendix~\ref{app:mitigation-methods}.

%\paragraph{Standout methods.}
\textit{Weighted Summaries} is a chunk-based approach that ensures balanced representation of the source text. The algorithm first splits the document into three segments (beginning, middle, and end). It then generates a partial summary for each segment independently, constraining the output length for each part according to a pre-allocated token budget (33\%, 34\%, and 33\% of the total desired summary length). These partial summaries are then combined to form the final output.

In \textit{Mirostat Decoding}, the model's temperature is dynamically adjusted during generation via a feedback loop (formula in Appendix Table~\ref{tab:mitigate_all}). This ensures the output text maintains a consistent level of unpredictability, which prevents the model from focusing too narrowly on initial sentences and encourages broader coverage of the source text. We use a target entropy of \(\mu^\star\!=\!2.0\) in our runs, similar to \citet{basu2021mirostatneuraltextdecoding}.

\subsection{Framing Effect Mitigation}
\textit{Weighted Token Decoding} directly
influences the model's token selection during the generation process. At each step, we modify the output probabilities by adding a logarithmic weight to the logits of specific tokens before sampling \cite{liu2021dexpertsdecodingtimecontrolledtext,dathathri2020plugplaylanguagemodels}. In particular, we down-weight tokens associated with negative sentiment to reduce framing shifts, as detailed in Appendix~\ref{app:mitigation-methods}. 

% \textcolor{violet}{We target negative lexemes because our framing transition analysis (Table~\ref{tab:framing_transitions_only}) shows that the most frequent shifts involve neutral contexts becoming negative or vice-versa, making negative sentiment the primary source of harmful framing changes.}

% A lighter alternative is the Self-Awareness prompt which yields comparable gains without modifying decoding.

\subsection{Hallucination Mitigation}

We analyse \textit{Knowledge Boundary Awareness}, which explicitly defines the model’s knowledge cutoff date in the context to establish clear temporal limits on what the model can reasonably know.
%, helping to prevent the fabrication of information about events beyond its training data by embedding these boundaries in prompts.\\ 
\textit{Epistemic Tagging} requires models to express confidence
levels alongside factual assertions, forcing self-evaluation of fact-checking abilities (i.e., the model additionally responds with a high or low confidence level for each response) \cite{lin2022teaching}.\\
Hallucination mitigation strategies are detailed in
Appendix~\ref{app:mitigation-methods} and prompt details provided in Appendix~\ref{app:prompt-strategies}.

\begin{table*}[h]
\small
\centering
\setlength{\tabcolsep}{3pt}
\begin{tabular}{lccccccccccccc}
\toprule
& \multicolumn{6}{c}{\textbf{Amazon Reviews}} & \multicolumn{6}{c}{\textbf{MediaSum News}} \\
\cmidrule(lr){2-7}\cmidrule(lr){8-13}
\textbf{Method} &
$\fphi$ \(\downarrow\) &
\(\bar{s}_b \uparrow\) & \(\bar{s}_m \uparrow\) & \(\bar{s}_e \uparrow\) &
\(\fpsi \downarrow\) & &
$\fphi$ \(\downarrow\) &
\(\bar{s}_b \uparrow\) & \(\bar{s}_m \uparrow\) & \(\bar{s}_e \uparrow\) &
\(\fpsi \downarrow\) &  \\
\midrule
\textbf{GPT-3.5-turbo} 
& 16.0\% &  0.848 & 0.826 & 0.825 & 7.6\% & 
& 24.8\% & 0.840 & 0.823 & 0.820 &6.1\%& \\
\textbf{Llama-3.2-3B-Instruct} 
& 14.9\% & 0.860 & 0.842 & 0.840 & 7.4\% & 
& 22.1\% & 0.851 & 0.837 & 0.832 &5.1\%& \\
\textbf{Llama-3-8B-Instruct} 
& 14.5\% & 0.855 & 0.837 & 0.836 & 7.0\%&
& 21.9\% & 0.847 & 0.834 & 0.828 & 4.0\%& \\
\textbf{Phi-3-mini-4k-Instruct} 
& 34.5\% & 0.836 & 0.822 & 0.823 &5.4\%& 
& 26.2\% & 0.828 & 0.827 & 0.823 & 4.9\%& \\
\textbf{Qwen3-4B-Instruct} 
& 27.8\% & 0.921 & 0.896 & 0.893 & 19.6\% &
& 45.9\% & 0.898 & 0.879 & 0.869 & 13.0\% & \\
\textbf{Gemma-3-27B-IT}
& 26.5\% & 0.924 & 0.896 & 0.892 & 23.7\% &
& 41.9\% & 0.902 & 0.879 & 0.868 & 17.6\% & \\
\bottomrule
\end{tabular}
\caption{
Models introduce framing changes ($\fphi$) and filter content based on position ($\fpsi$).
$\fphi$ is the proportion of examples where the framing changed from the original to the summary (lower is better). 
\(\bar{s}_b\), \(\bar{s}_m\), and \(\bar{s}_e\) represent the average cosine similarity between the summary and the beginning, middle, and end thirds of the source text, respectively (higher is better). 
$\fpsi$ is the percentage of examples with primacy bias as defined in Equation~\ref{eq:primacy-bias-threshold} (lower is better). All differences between \(\bar{s}_b\) and \(\bar{s}_m\) are statistically significant ($p < .001$).
}
\label{tab:framing-primacy-bias-results}
\end{table*}

%\section{Experiments}

\section{Results}
\subsection{LLMs Alter Framing During Summarization}
\label{sec:framing_primacy_results}

\begin{table*}[h]
\centering
\small
\setlength{\tabcolsep}{2.3pt}
\begin{tabular}{llcccccc}
\toprule
\textbf{Dataset} & \textbf{Model} & Neu$\rightarrow$Pos & Neu$\rightarrow$Neg & Pos$\rightarrow$Neg & Neg$\rightarrow$Pos & Pos$\rightarrow$Neu & Neg$\rightarrow$Neu \\
\midrule
\multirow{6}{*}{MediaSum} 
& \textbf{GPT-3.5-turbo}                 & 2.9\% & 8.6\% & 0.1\% & 0.1\% & 1.9\% & 11.2\% \\
&\textbf{Llama-3-8B-Instruct}    & 2.6\% & 8.0\% & 0.0\% & 0.1\% & 2.3\% & 8.9\% \\
& \textbf{Llama-3.2-3B-Instruct}  & 2.0\% & 7.7\% & 0.1\% & 0.1\% & 2.3\% & 9.9\% \\
& \textbf{Phi-3-mini-4k-Instruct}         & 2.6\% & 7.3\% & 0.1\% & 0.7\% & 2.4\% & 13.1\% \\
& \textbf{Qwen3-4B-Instruct}              & 7.6\% & 31.4\% & 0.7\% & 0.3\% & 2.0\% & 2.9\% \\
& \textbf{Gemma-3-27B-IT}                 & 5.2\% & 30.8\% & 0.8\% & 0.1\% & 1.6\% & 3.0\% \\
\midrule
\multirow{6}{*}{Amazon} 
& \textbf{GPT-3.5-turbo}                 & 2.6\% & 0.5\% & 0.2\% & 1.4\% & 7.4\% & 3.9\% \\
& \textbf{Llama-3-8B-Instruct}    & 2.3\% & 0.6\% & 0.7\% & 1.4\% & 6.7\% & 2.8\% \\
& \textbf{Llama-3.2-3B-Instruct}  & 1.8\% & 0.6\% & 0.5\% & 2.1\% & 6.2\% & 3.7\% \\
& \textbf{Phi-3-mini-4k-Instruct}          & 1.6\% & 0.8\% & 1.8\% & 2.2\% & 21.2\% & 6.9\% \\
& \textbf{Qwen3-4B-Instruct}              & 15.3\% & 3.9\% & 1.5\% & 1.4\% & 1.5\% & 1.3\% \\
& \textbf{Gemma-3-27B-IT}                 & 14.4\% & 4.2\% & 1.8\% & 1.3\% & 1.2\% & 0.7\% \\
\bottomrule
\end{tabular}
\caption{Framing category transitions in MediaSum and Amazon Reviews datasets. Values represent the percentage of summaries transitioning from framing category 'x' to 'y' (x$\rightarrow$y).}
\label{tab:framing_transitions_only}
\end{table*}
We observe that models alter the framing of summaries, compared to the source content, in both media interviews and the Amazon dataset (Table~\ref{tab:framing-primacy-bias-results}). GPT-3.5-turbo and Llama models exhibit a framing change fraction between 14.5\% and 16\%, whereas Gemma-3-2-27B-IT, Qwen3-4B-Instruct, and Phi-3-mini-4k-Instruct show framing changes between 26.5\%  and 45.9\%. 
%All aggregate statistics reported in the abstract are computed using a mean-of-means approach across all six models (GPT-3.5-turbo, Llama-3-8B-Instruct, Llama-3.2-3B-Instruct, Phi-3-mini-4k-Instruct, Qwen3-4B-Instruct, Gemma-3-27B-IT) and both datasets (Amazon Reviews and MediaSum News), where each model-dataset pair contributes one mean calculated from 1000 examples.
In Table~\ref{tab:examples}, we show an example of a framing change between a summary and the ground-truth context.
Table~\ref{tab:framing_transitions_only} provides a detailed evaluation of framing alterations (e.g., from neutral to positive, or positive to negative). In the MediaSum news interviews, most framing shifts occur between neutral and negative, ranging from 7.3\% to 8.6\% for GPT-3.5-turbo and Llama models, and reaching 30.8\% to 31.4\% for Qwen3-4B-Instruct and Gemma-3-27B-IT. Shifts from negative to neutral range from 2.9\% to 13.1\% across models.
%, which suggests that models often introduce negative sentiment. 
In Amazon reviews, the most common shift is from positive to neutral, ranging from 6.2\% to 7.4\% for the Llama and GPT-3.5-turbo models, and 21.2\% for Phi-3-mini-4k-Instruct. Qwen3-4B-Instruct and Gemma-3-27B-IT show higher rates of neutral-to-positive shifts (14.4\% to 15.3\%) compared to other models, suggesting different framing patterns across model families.

\begin{table*}[h]
\centering
\small
\footnotesize
\setlength{\tabcolsep}{3.5pt}
\begin{tabularx}{\linewidth}{p{0.77\linewidth}|p{0.2\linewidth}}
\toprule
\textbf{Original Review} & \textbf{Summary} \\
\midrule
``This is a fun, useful tablet for the price. I love that you can make all different homepages. Love that you can pick wallpaper which you can't do with Kindle Fires, which are the only other tablets I've owned. Of course, if you want a tablet mainly for books, Kindle is the tablet you want. I also love all the apps you can get with the Samsung.

\par What I don't like is the 8\,GB storage. I was under the impression that if you got the 32\,GB extra, it would integrate with the tablet rather than just be like a drive for MP3s and such. 8 gigs is barely enough for the apps you need. Also, the manual tells you nothing about how to add things to the extra drive. Had to go to YouTube.

\par My tablet has connection issues from time to time. I still really enjoy it most of the time. It can be super fast and plays things well from the cloud. Very light and attractive.

\par I am returning the tablet. 8\,GB is just not enough. Extra storage is cumbersome to use and doesn't work half of the time. Not easy like the cloud. Speakers unimpressive. I am disappointed because I thought I would like the Samsung. Also a battery hog. Will try the Nexus 7.'' &
``I enjoyed the Samsung tablet for its customization options and app variety, but the 8 GB storage was insufficient. The extra storage was cumbersome and unreliable. Connection issues and battery drain were also drawbacks. I will be returning it and trying the Nexus 7 instead.'' \\
\bottomrule
\end{tabularx}
\caption{Example of framing shift: from neutral (original text) to negative (summary by GPT-3.5-turbo).}
\label{tab:examples}
\end{table*}

% \begin{table*}[t]
% \centering
% \scriptsize
% \setlength{\tabcolsep}{5pt}
% \begin{tabular}{llccccccc}
% \toprule
% \textbf{Dataset} & \textbf{Model} & \textbf{$\fphi$ } & Neu~$\rightarrow$~Pos & Neu~$\rightarrow$~Neg & Pos~$\rightarrow$~Neg & Neg~$\rightarrow$~Pos & Pos~$\rightarrow$~Neu & Neg~$\rightarrow$~Neu \\
% \midrule
% \multirow{4}{*}{MediaSum} 
% & GPT-3.5-turbo                     & 24.8\% & 2.9\% & 8.6\% & 0.1\% & 0.1\% & 1.9\% & 11.2\% \\
% & Llama-3-8B-Instruct        & 21.9\% & 2.6\% & 8.0\% & 0.0\% & 0.1\% & 2.3\% & 8.9\% \\
% & Llama-3.2-3B-Instruct      & 22.1\% & 2.0\% & 7.7\% & 0.1\% & 0.1\% & 2.3\% & 9.9\% \\
% & Phi-3-mini-4k-Instruct    & 26.2\% & 2.6\% & 7.3\% & 0.1\% & 0.7\% & 2.4\% & 13.1\% \\
% \midrule
% \multirow{4}{*}{Amazon} 
% & GPT-3.5-turbo                     & 16.0\% & 2.6\% & 0.5\% & 0.2\% & 1.4\% & 7.4\% & 3.9\% \\
% & Llama-3-8B-Instruct        & 14.5\% & 2.3\% & 0.6\% & 0.7\% & 1.4\% & 6.7\% & 2.8\% \\
% & Llama-3.2-3B-Instruct      & 14.9\% & 1.8\% & 0.6\% & 0.5\% & 2.1\% & 6.2\% & 3.7\% \\
% & Phi-3-mini-4k-Instruct    & 34.5\% & 1.6\% & 0.8\% & 1.8\% & 2.2\% & 21.2\% & 6.9\% \\
% \bottomrule
% \end{tabular}
% \caption{\textcolor{ForestGreen}{Framing transition percentages across the MediaSum and Amazon Reviews datasets (context length 50-4000 tokens). Each x$\rightarrow$y refers to the percentage of summaries that shifted from framing category x to y.}}
% \label{tab:framing_transitions_combined1}
% \end{table*}

\subsection{LLM Summaries have Imbalanced Context Coverage}
We find that, on average, summaries align more closely with the beginning of the source text than with the middle and end sections (Table~\ref{tab:framing-primacy-bias-results}). 
For example, across all models and datasets, the average similarity to the beginning segment is consistently higher than the similarity to the middle and end segments. Paired t-tests between the beginning-summary similarity ($s_b$) and middle-summary similarity ($s_m$) are statistically significant across all evaluated models and datasets ($p < .001$). The detailed test statistics are provided in Appendix~\ref{app:stat-tests}.
We further quantify this coverage imbalance using our primacy bias score. For the Amazon Reviews dataset, $\fpsi$ ranges from 5.4\% to 23.7\%. Similarly, MediaSum summaries exhibit $\fpsi$  scores between 4.0\% and 17.6\%, which shows primacy bias in structured news content.

\subsection{LLMs Hallucinate on Post-Knowledge-Cutoff Data}
Table~\ref{tab:llm_hallucination} shows a consistent decrease in strict accuracy $(s_{H}^T)$ from pre-cutoff to post-cutoff data across all models. Llama-3-8B-Instruct declined from 26\% to 21\% ($\delta_{H}=5\%$), Llama-3.2-3B-Instruct dropped from 19\% to 13\% ($\delta_{H}=6\%$), and Phi-3-mini-4k-Instruct showed a $\delta_{H}=4\%$ decline.\\
%While the 
%modest 
%difference in strict accuracy between pre- and post-cutoff data suggests that a model’s performance does not deteriorate dramatically when encountering newer events,
The consistently low strict accuracy, compared to actual news and falsified news accuracy, highlights a critical limitation: %Even when models achieve relatively high accuracies on true and falsified news separately, their low strict accuracy reveals 
\textit{the persistent inability to reliably differentiate fact from fabrication}. Strong performance on isolated real or falsified news does not guarantee that models can reliably distinguish truth from falsehood.% when presented together.

\begin{table*}[t]
\centering
\small
\scriptsize
\setlength{\tabcolsep}{3pt}
\renewcommand{\arraystretch}{1.1}

\begin{tabularx}{\textwidth}{@{}
  p{2.6cm}                       
  p{3.0cm}
  *{6}{>{\centering\arraybackslash}X}
  @{}}
\toprule
\multirow{2}{*}{\textbf{Models}} &
\multirow{2}{*}{\textbf{Prompt Strategy}} &
\multicolumn{3}{c}{\textbf{Pre-cutoff data}} &
\multicolumn{3}{c}{\textbf{Post-cutoff data}} \\
\cmidrule(lr){3-5}\cmidrule(l){6-8}
& &
\makecell{Actual News\\Acc.} &
\makecell{Falsified News\\Acc.} &
\makecell{Strict Acc.\\$(s_{H}^T)$} &
\makecell{Actual News\\Acc.} &
\makecell{Falsified News\\Acc.} &
\makecell{Strict Acc.\\$(s_{H}^T)$} \\
\midrule
\multirow{4}{2.6cm}{\textbf{Llama-3-8B-Instruct}}
    & Baseline                   & 74\% & 46\% & 26\% & 75\% & 42\% & 21\% \\
    & Prompt Calibration (CoT)   & 39\% & 85\% & \textbf{27\%} & 36\% & 84\% & \textbf{22\%} \\
    & \makecell[l]{Knowledge Boundary Aware} & 65\% & 54\% & 25\% & 50\% & 63\% & 18\% \\
    & Epistemic Tagging          & 80\% & 35\% & 18\% & 78\% & 37\% & 19\% \\
\midrule
\multirow{4}{2.6cm}{\textbf{Llama-3.2-3B-Instruct}}
    & Baseline                   & 48\% & 58\% & 19\% & 35\% & 72\% & 15\% \\
    & Prompt Calibration (CoT)   & 48\% & 56\% & 17\% & 46\% & 63\% & 15\% \\
    & \makecell[l]{Knowledge Boundary Aware} & 25\% & 87\% & 15\% & 15\% & 94\% & 10\% \\
    & Epistemic Tagging          & 51\% & 68\% & \textbf{26\%} & 53\% & 64\% & \textbf{23\%} \\
\midrule
\multirow{4}{2.6cm}{\textbf{Phi-3-mini-4k-Instruct}}
    & Baseline                   & 14\% & 97\% & 12\% & 9\% & 98\% & 8\% \\
    & Prompt Calibration (CoT)   & 3\%  & 99\% & 3\%  & 3\% & 99\% & 3\% \\
    & \makecell[l]{Knowledge Boundary Aware} & 3\%  & 99\% & 3\%  & 2\% & 99\% & 2\% \\
    & Epistemic Tagging          & 45\% & 76\% & \textbf{31\%} & 53\% & 70\% & \textbf{29\%} \\
\bottomrule
\end{tabularx}

\caption{Accuracies on real news items and their falsified counterparts, evaluated before and after each model’s training cutoff date.  Strict Accuracy $s_{H}^T(D_j)$ counts predictions where the model gets both members of each (True, False) pair correct.}
\label{tab:llm_hallucination}
\end{table*}

\subsection{Special Biases Require Special Mitigations}
\label{sec:insights}

We discuss trade-offs among mitigation methods for different biases, and how their effectiveness varies depending on the model and corpus. For example, techniques aimed at reducing position-based salience (e.g., primacy bias) often work by redistributing attention across the input. However, this can disrupt the model's ability to maintain consistent sentiment in its responses. All definitions of the 18 mitigation approaches we tested are in Appendix~\ref{app:mitigation-methods}.

\begin{table}[t]
\centering\footnotesize
\setlength{\tabcolsep}{3pt} % Adjusted for better spacing
\begin{tabular}{lccc}
\toprule
\textbf{Method} &
$\bar{s}$$^{\dagger}$ &
$\fpsi$\% $^{\ddagger}$ &
$\fphi$\% $^{\S}$\\
\midrule
\textbf{Baseline} & 0.843 & \textbf{7.0\%} & 14.5\% \\
\midrule
Weighted Summaries      & \textbf{0.910} & 15.1\% & 15.1\% \\
Mirostat Decoding       & 0.902 & 15.5\% & 13.8\% \\
Pos.-Invariant Shuffle  & 0.891 & 11.5\% & 21.2\% \\
Weighted Token Decoding & 0.858 & 18.4\% & \textbf{13.6\%} \\
Self-Awareness Prompt   & 0.889 & 22.9\% & 15.2\% \\
\bottomrule
\end{tabular}
\caption{Key mitigation results on Amazon Reviews for Llama-3-8B-Instruct.
$^{\dagger}$Mean content-coverage similarity $(\bar{s}_b+\bar{s}_m+\bar{s}_e)/3$.
$^{\ddagger}$Primacy-bias score.
$^{\S}$Framing-change rate. Lower is better for $\fpsi$ and $\fphi$. Behavior on other corpora/models follows a similar qualitative pattern; see full results in Table~\ref{tab:combined}.}
\label{tab:mini_results}
\vspace{-1em}
\end{table}

\subsubsection{Primacy Effect and Coverage}
\label{sec:insights_primacy}
\noindent The \textit{\textbf{Weighted Summaries}} method, which assigns a fixed token budget to each text segment, consistently increases overall content coverage. For instance, with Llama-3-8B-Instruct on Amazon Reviews (Table~\ref{tab:mini_results}), it increases average coverage similarity from a baseline of 0.843 to 0.910. However, this comes at the cost of an increased primacy bias score ($\fpsi$), which increases from 7.0\% to 15.1\%.

\noindent\textit{\textbf{Mirostat Decoding}} also improves coverage (Table~\ref{tab:combined}); it is the \emph{only} method that reduces $\fpsi$ (primacy bias) on the smaller Phi-3 model by 1.2 percentage points (from 5.4\% to 4.2\%) but worsens it on larger models (e.g., from 7.0\% to 15.5\% on Llama-3-8B for Amazon).

\noindent\textit{\textbf{Position-Invariant Shuffle}} is useful diagnostically because it helps confirm that the model relies on word order. By randomly shuffling sentences, we remove positional cues; the resulting change in output reveals the model's sensitivity to input structure. While this method increases overall coverage (e.g., from 0.843 to 0.891 for Llama-3-8B on Amazon Reviews; Table~\ref{tab:combined}), it degrades other metrics, increasing the primacy score $\fpsi$ to 11.5\% and the framing change rate $\fphi$ to 21.2\%, and comes at the risk of losing temporal information.

\subsubsection{Framing Effect and Consistency}
\label{sec:insights_framing} 

\textit{\textbf{Weighted Token Decoding}} is the most effective method for mitigating framing bias. By downweighting negative lexemes during decoding, it achieves the lowest framing-change fraction ($\fphi$) for Llama-3-8B, reducing it by 0.9 percentage points to 13.6\% (Table~\ref{tab:combined}). However, this targeted intervention significantly worsens primacy bias, with $\fpsi$ increasing from 7.0\% to 18.4\%.

\noindent\textit{\textbf{Prompt-only approaches:}}
Prompts such as \textbf{Self-Awareness} or \textbf{Chain-of-Thought}, alter $\fphi$ by only a few percentage points in most settings. However, the Self-Awareness variant yields the best framing result on MediaSum for Llama-3-8B-Instruct, reducing $\fphi$ to 21.0\% from a 21.9\% (Table~\ref{tab:combined}) baseline at a negligible computational cost.

\subsubsection{Prompt Calibration and Epistemic Tagging improve Hallucination}
% \textcolor{red}{The section is too long and frames the small improvements (e.g., 1\% in strict accuracy) as significant. \textcolor{blue}{We can make this concise in that case indicating that this general mitigation strategy resulted in a small improvement of 1\% as a head-start for other mitigation strategies. Update: reduced the content}}

%\textit{\textbf{Prompt Calibration using Chain-of-Thought.}} 
%This mitigation strategy improves strict accuracy for Llama-3-8B-Instruct by 1\% on both pre and post cut-off data. \\
%For Llama-3.2-3B-Instruct, strict accuracy drops by 2\% on pre-cutoff data and there is no significant improvement when applied on post cut-off data. \\
%Phi-3-mini-4k-Instruct showes the most dramatic degradation. Real news accuracy plummeted from 14\% to 3\% on pre-cutoff data. Post-cutoff data showed similar degradation in real news detection while maintaining high falsified news detection.
%The approach's marginal improvement for the larger Llama-3-8B-Instruct model is overshadowed by the significant performance degradation it causes in smaller models like Phi-3-mini-4k-Instruct, consistent with known limitations in factual recall for smaller language models \cite{abdin2024phi3technicalreporthighly}.

% Prompt Calibration using Chain-of-Thought showed limited utility, as it marginally improved accuracy on Llama-3-8B-Instruct, but increased error on all others.
Applying \textbf{Prompt Calibration with Chain-of-Thought} enhances strict accuracy for Llama-3-8B Instruct but shows limited utility for others. It improves the detection of falsified news by 40\% for both pre- and post-cut-off data.

\textbf{Knowledge Boundary Awareness} improves Llama-3-8B-Instruct's performance by 8\% and 21\% in falsified news detection in pre and post cut-off data, respectively, alongside maintaining its strict accuracy close to the baseline. For the other models, performance declines, despite observing a significant improvement in the detection of falsified news data.

% \noindent\textbf{Epistemic Tagging} performs best across all models. It associates a confidence level during the fact-checking process (i.e., high or low) with its response about whether an event occurred (Table \ref{tab:confidence_scores_ablation_study}) \\
% We observe an increase in strict accuracy for Phi-3-mini-4k-Instruct by 21\% on post cut-off data and by 19\% on pre cut-off data. The Llama-3.2-3B-Instruct also shows improved accuracy by 8\% on post-cut-off data and by 7\% on pre-cut-off data, along with balanced independent accuracies for real and falsified news data above 50\%.

\noindent\textbf{Epistemic Tagging} performs best across all models. It associates a confidence level during the fact-checking process (i.e., high or low) with its response about whether an event occurred (Table \ref{tab:confidence_scores_ablation_study}) \\
We observe an increase in strict accuracy for Phi-3-mini-4k-Instruct by 21\% on post cut-off data and by 19\% on pre cut-off data. The Llama-3.2-3B-Instruct also shows improved accuracy of 8\% on post-cut-off data and 7\% on pre-cut-off data, with balanced independent accuracies above 50\% for real and falsified news data.

\begin{table}[t]
\centering
\small
\scriptsize
\setlength{\tabcolsep}{4pt}
\begin{tabular}{llrrrr}
\toprule
\textbf{Models} &
\textbf{cutoff}& 
\multicolumn{2}{c}{\textbf{Actual News}} & 
\multicolumn{2}{c}{\textbf{Falsified News}} \\
\cmidrule(lr){3-4} \cmidrule(lr){5-6}
& & \textbf{High} & \textbf{Low} & \textbf{High} & \textbf{Low} \\
\midrule
\multirow{2}{*}{\textbf{Llama-3-8B-Instruct}} 
    & Pre  & 99.0\% & 1.0\% & 99.8\% & 0.2\% \\
    & Post & 98.3\% & 1.7\% & 99.9\% & 0.1\% \\
\midrule
\multirow{2}{*}{\textbf{Llama-3.2-3B-Instruct}} 
    & Pre  & 75.8\% & 24.2\% & 24.6\% & 75.4\% \\
    & Post & 66.7\% & 33.3\% & 83.3\% & 16.7\% \\
\midrule
\multirow{2}{*}{\textbf{Phi-3-mini-4k-Instruct}} 
    & Pre  & 93.8\% & 6.2\% & 86.2\% & 13.8\% \\
    & Post & 98.1\% & 1.9\% & 95.4\% & 4.6\% \\
\bottomrule
\end{tabular}
\caption{Epistemic Tagging prediction Confidence levels (high, low) for different models across cutoff conditions.}
\label{tab:confidence_scores_ablation_study}
\end{table}

\section{Impact on Human Decision-Making}
When LLMs modify the framing or emphasize certain aspects of the content they process, they can inadvertently shape human perception and decision-making when humans are exposed to the content. In this section, we empirically demonstrate that human decisions change when LLMs process the original text and introduce undesired changes to it.

\paragraph{Experiment Design} Participants choose between two manufacturers of a product (e.g., a headlamp) based on product reviews. We probe 1000 reviews for items in the home, electronics, and kitchen categories from the \texttt{Amazon-Reviews-2023} dataset \cite{hou2024bridging} and generate summaries using GPT-3.5. To analyze the worst-case scenario of how LLM
content alteration can influence decision-making, we use GPT-4 to identify product types with the most extreme framing changes and manually select
products to show to participants. Specifically, we select 10 product pairs where one product review shifts from negative/neutral to positive after summarization, while the other maintains neutral framing. 

Participants are divided into two conditions and view 10 product pairs, either as original reviews or summaries (5 of each format per participant, alternating between conditions). We measure manufacturer selection rates based on the framing of reviews.
Additionally, participants are asked to provide their Willingness to Pay \cite{mitchell2013using} (WTP) for the product from the manufacturer they picked. WTP describes the maximum price someone would pay, knowing the cost of another product. For example, if the participant selects manufacturer $M$ whose product is described by a positive summary, they learn that the product from manufacturer $W$ costs $\$X_W$ and decide how much more over $\$X_W$ they would pay to $M$. This measure enables us to quantify the strength of the change in behavior induced by altering the content by the LLM. 

\paragraph{Hypothesis} 
We hypothesize that participants will be more likely to choose a manufacturer when they read a positively framed LLM summary of their product compared to when they read the original review (which is neutral or negative). Additionally, we posit that their Willingness to Pay will be higher when reading the positively framed summaries.

\paragraph{Participants} We recruit 72 participants on Prolific (32-40 per condition, English-speaking, 33 men, 39 women, aged 19-75). Participants are presented with instructions about the study and take a short comprehension quiz at the beginning of the study, before proceeding to the decision-making stage. Based on this quiz, we accept 70 submissions (3\% rejection rate). On average, the experiment takes 32 minutes and pays \$4.83, which includes \$3 base pay and a performance bonus for selecting the option that matches the majority opinion (encouraging participants to make decisions they believe reflect typical consumer preferences).

\begin{figure}
    \centering
    \includegraphics[width=0.99\linewidth, page=1]{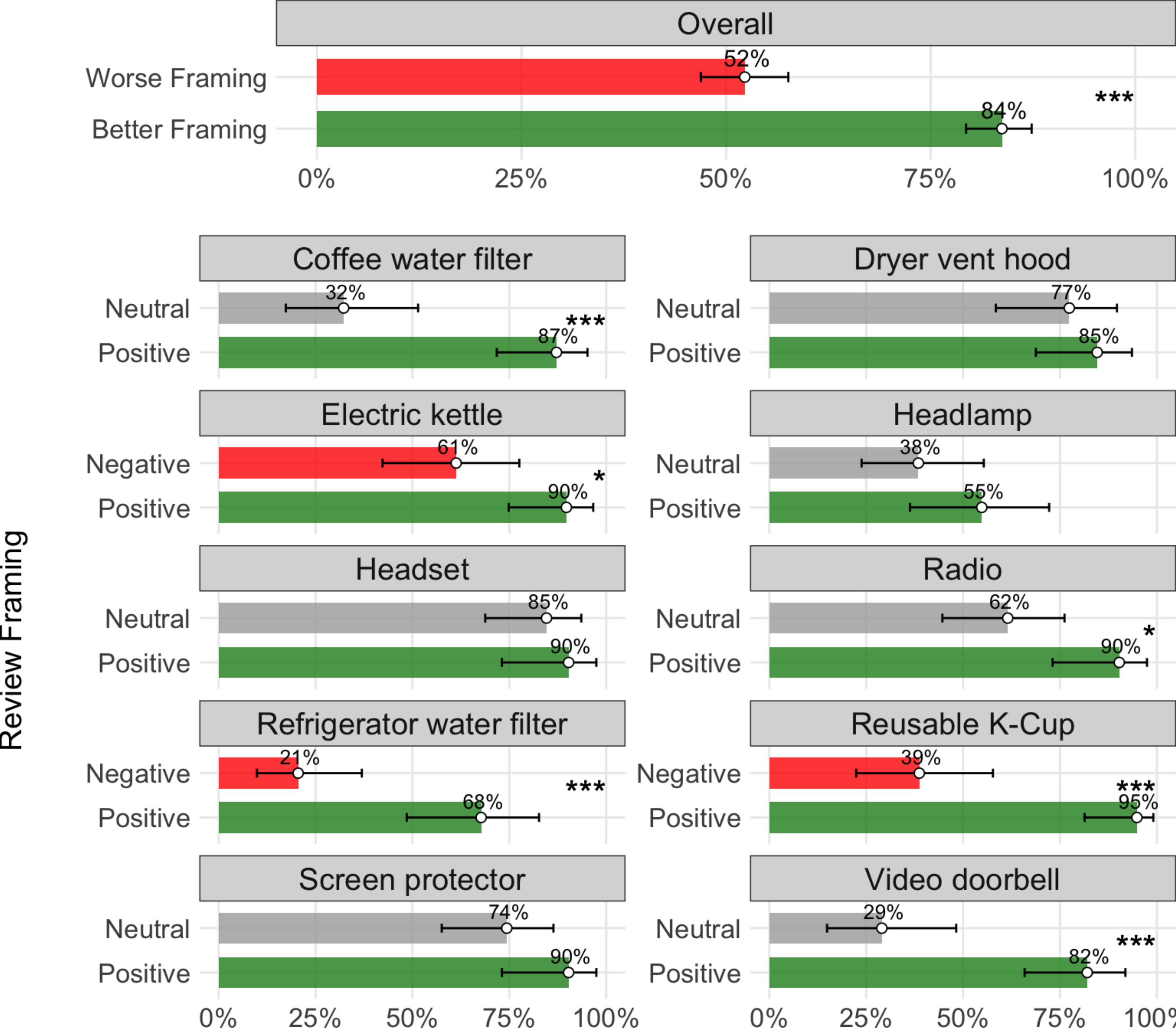}
    \caption{Mean selection rates for manufacturers depending on whether participants read the original review of their product (neutral or negative framing) or its summary (positive framing). Error bars display 95\% confidence intervals and stars denote significance levels (* p < 0.05, ** p < 0.01, *** p < 0.001).}
    \label{fig:ExpRes}
\end{figure}

\paragraph{Results} We see that humans make different decisions, depending on the framing of the text they are exposed to. Choice rates differ significantly between framing conditions ($\chi^2$ = 78, df = 1, $p < 0.001$). \cref{fig:ExpRes} shows that participants picked manufacturers when their products were described by summaries (positive framing) in 83.7\% of the cases compared to 52.3\% of the cases if these products were described by original reviews (neutral or negative framing). The effect is robust across products, but differs in strength (\cref{fig:ExpRes}). Additionally, we find that reading the better-framed summaries increases participants' willingness to pay by 4.5\% with respect to the original product price ($t = 4.406, p < 0.001$ according to the linear mixed-effects model). \textit{Together, these results demonstrate that exposing participants to summarized content significantly affects whether they would purchase a product and what price they would pay for it.} This demonstrates that when LLM summarization alters the text's framing, it can have measurable consequences on human behavior.

\section{Conclusion}
This paper quantifies content changes in summarization and news fact-checking tasks performed by LLMs. 
%We find that models significantly alter content. 
Averaged across all tested models, models introduce framing bias to the user in 26.42\% of instances, and show primacy bias in 10.12\% of the cases, where summaries disproportionately reflect content from the beginning of the original text. Users are exposed to hallucinations on post-knowledge-cutoff content in 60.33\% of instances ( averaged across all tested models). Through our user study, we find that model bias (specifically, framing bias that occurs in review summarization) influences people's decisions. We observe that a positive framing bias increases the likelihood that people will purchase a given product and enhances their perception of its value.

%We evaluate 18 mitigation strategies, including both specific strategies tailored to each bias category and general-purpose strategies. 
By evaluating 18 mitigation strategies, we find that each approach's effectiveness depends on the model and the targeted bias. For example, \textit{Weighted Summaries} and \textit{Mirostat Decoding} show reduced framing changes and primacy biases, particularly in smaller and mid-sized models. \textit{Epistemic Tagging} consistently improved factual reliability regarding post-knowledge-cutoff data for smaller models.

Our paper represents a step toward careful analysis and mitigation of content alteration introduced by LLMs to humans, and provides insight into its effects, aiming to reduce the risk of systemic bias in decision-making across media, education, and public policy.

\section{Limitations}

This study investigates the degree to which LLM outputs modify original content in ways that may impact human judgment. Most of our mitigation methods were designed with one specific issue in mind (e.g., framing, hallucinations, or primacy effects). In our future work, we intend to explore more general mitigation techniques that address overlapping biases simultaneously. 

Our current primacy bias score uses a fixed, generic threshold to identify summaries that disproportionately focus on the beginning of the source text. While this threshold provides a consistent way to detect bias, our future work aims to explore ways to refine it.  In addition, our evaluation relies on one LLM to assess another LLM's outputs in the framing bias metric. Although commonly used, LLM-as-a-judge approaches have drawbacks. We use LLM-as-a-judge to the best of our knowledge, given current best practices.
%, which can introduce systematic errors. Using human evaluators remains important.

Our work aims to encourage the NLP community to continuously test LLMs for biases that can influence users' decision-making, which is why we introduce the self-updating NewsLensSync dataset. %\cite{news_lens_sync_contribution}. Currently, the dataset holds articles related to the "political" domain. In our future work, we aim to incorporate other areas to make it more generalizable and usable across a diverse set of researchers. %However, our data could be misused for training models to reinforce existing ideological narratives or use falsified data. 
We urge researchers to exercise caution when framing and contextualizing their use of the data. We provide our data under the CC-BY-4.0 license.\\

\textbf{Ethical Considerations.} 

Our analysis reveals that LLMs tend to focus more on content from the beginning of the context, exhibit hallucination of facts, and alter the text framing in their outputs. These alterations can affect the user and influence their decision-making. These findings highlight the risks associated with working with LLMs, even in everyday tasks. 
\newline

%specifically when used in high-stakes applications. 
%In future work, we aim to examine the reasoning behind such errors in the model outputs and analyse how such biases affect users through a user study. \\
% \newline
%\textbf{Data.} Our work aims to encourage the NLP community to continuously test LLMs for biases that can influence users' decision-making, which is why we introduce the self-updating NewsLensSync dataset \cite{news_lens_sync_contribution}.
%that utilizes NewsAPI \cite{newsapi_getstarted_2025} to create a falsified version of the news description and news content using the Negate python package used in previous work \cite{anschutz2023negbleurt}. 

% \newline
\textbf{Experiments. } All experiments are run on NVIDIA RTX A6000  for open-source models and the official APIs for closed-source models with a fixed random seed. Our run time was approximately 360 GPU hours. All prompts details are included in Appendix~\ref{app:prompt-strategies}.

\bibliography{custom}

\clearpage

\appendix

\section{Appendix}
\subsection{LLM as a framing judge}
\label{app:sentiment-classifier}
\renewcommand{\thetable}{A\arabic{table}}
\setcounter{table}{0}
Previous work has shown that high-capability LLMs like GPT-4 can achieve over 80\% agreement with human experts on question-answering tasks, with the level of agreement observed among humans themselves being 81\% \ cite {zheng2023judging}. GPT-4 reached 85\% agreement with human judgments, suggesting that LLMs can serve as effective proxies for human preferences \cite{zheng2023judging}. Thus, we test three high-capability models to classify the framing in our study ( GPT-3.5-turbo, GPT-4-turbo, and GPT-o4-mini). 
On a sample of 500 randomly selected Amazon reviews, we use the user-provided ratings as ground truth to evaluate the models' accuracy. We ask the models to rate the product based on the review, then we map ratings 1 and 2 as negative, 3 as neutral, and 4 and 5 as positive. The GPT-3.5-turbo, GPT-4-turbo, and GPT-o4-mini achieve accuracy of 0.89, 0.91, and 0.92, respectively. Based on this, we select GPT-o4-mini to serve as the framing judge. We prompt the model as follows:
\begin{quote}
    \small ``[TEXT]\\
    Classify the framing of the text as Positive, Negative, or Neutral. Respond with the class label only.''
\end{quote}to obtain the text framing.

\subsection{Model Setup and Hyperparameters}\label{app:hyperparams}

% Our implementation utilizes the Hugging Face Transformers library using the official APIs. We employ Flash Attention 2 when available to enhance computational efficiency during inference. For consistency across experimental conditions, we maintain fixed generation parameters with \texttt{temperature=0.01}, \texttt{do\_sample=False}, and \texttt{max\_new\_tokens= 500} set to a constant with fixed random seed. Table \ref{tab:models} shows the models we evaluate, their context window, and the rationale behind choosing them. 

% All aggregate statistics reported in the abstract are computed using a mean-of-means approach across all six models (GPT-3.5-turbo, Llama-3-8B-Instruct, Llama-3.2-3B-Instruct, Phi-3-mini-4k-Instruct, Qwen3-4B-Instruct, Gemma-3-27B-IT) and both datasets (Amazon Reviews and MediaSum News), where each model-dataset pair contributes one mean calculated from 1000 examples.

Our implementation uses the Hugging Face Transformers library via its official APIs. We employ Flash Attention 2 when available to enhance computational efficiency during inference. For consistency across experimental conditions, we maintain fixed generation parameters with \texttt{temperature=0.01}, \texttt{do\_sample=False}, and \texttt{max\_new\_tokens= 500} set to a constant with a fixed random seed. Table \ref{tab:models} presents the models we evaluate, their context windows, and the rationale for choosing them. 

All aggregate statistics reported in the abstract are computed using a mean-of-means approach across all six models (GPT-3.5-turbo, Llama-3-8B-Instruct, Llama-3.2-3B-Instruct, Phi-3-mini-4k-Instruct, Qwen3-4B-Instruct, Gemma-3-27B-IT) and both datasets (Amazon Reviews and MediaSum News), where each model-dataset pair contributes one mean calculated from 1000 examples.

\begin{table}[h]
    \centering
    \scriptsize
    \begin{tabularx}{\linewidth}{@{}
        >{\raggedright\arraybackslash}p{2.3cm} 
        >{\centering\arraybackslash}p{0.9cm}
        X                              
        @{}}
        \toprule
        \textbf{Model} & \textbf{Context Window} & \textbf{Reason} \\
        \midrule
        GPT-3.5-turbo         & 16k & A closed-source, highly capable model that is cost-effective~\cite{openai2024gpt35turbo}. \\[2pt]
        Llama-3-8B-Instruct   & 8k  & Outperforms competing models on both per-category win rate and average per-category score~\cite{grattafiori2024llama3herdmodels}. \\[2pt]
        Llama-3.2-3B-Instruct & 8k  & Lightweight yet powerful; offers multilingual text-generation abilities~\cite{meta2024llama3}. \\[2pt]
        Phi-3-mini-4k-Instruct & 4k & Matches high-performance closed-source models on several key benchmarks~\cite{abdin2024phi3technicalreporthighly}.  \\[2pt]
        Qwen3-4B-Instruct & 256k & A small, high-performing open-source model included to broaden architectural diversity~\cite{yang2025qwen3technicalreport}. \\[2pt]
        Gemma-3-27B-IT & 128k & A large, state-of-the-art open-source model included to test for scale effects~\cite{gemmateam2025gemma3technicalreport}. \\
        \bottomrule
    \end{tabularx}
    \caption{Models used in our evaluation, along with their context-window sizes and selection rationale.}
    \label{tab:models}
\end{table}

Table \ref{Cutoffs} lists the cut-off for the models used for our hallucination study.

\begin{table}[h]
    \centering
    \scriptsize
    \begin{tabularx}{\linewidth}{@{}
        >{\raggedright\arraybackslash}p{4.5cm} 
        >{\centering\arraybackslash}X
        @{}}
        \toprule
        \textbf{Model} & \textbf{Cut-off date (training data freshness)} \\
        \midrule
        Llama3-8b-instruct & Mar 2023 \\
        Llama 3.2-3b-instruct & Oct 2024 \\
        Phi-3-mini-4k-instruct & Oct 2023 \\
        \bottomrule
    \end{tabularx}
    \caption{Models and their cut-off dates for training data freshness.}
    \label{Cutoffs}
\end{table}

\subsection{Further Information on our NewLensSync Dataset}\label{appendix:NewLensSync}\label{app:stats}
We use a fixed list of trusted domains: \texttt{bbc.co.uk}, \texttt{reuters.com}, \texttt{apnews.com}, \texttt{npr.org}, \texttt{pbs.org}, \texttt{theguardian.com}, \texttt{wsj.com}, \texttt{nytimes.com}, and \texttt{propublica.org}.

The NewsLensSync orchestrator builds a datastore for each article containing \texttt{source}, \texttt{author}, \texttt{title}, \texttt{api\_description} (from NewsAPI), \texttt{webscraped\_description} (extracted directly from the article to overcome the API limit), \texttt{falsified\_description} (negated version for hallucination testing), \texttt{api\_content}, \texttt{url}, and \texttt{publishedAt}. 

A scheduled CRON job performs one API call per day over the trusted list, minimizing duplication. Any rare overlaps are removed during preprocessing, retaining only unique records for LLM evaluation. Duplication checking is deferred to evaluation time to avoid overhead. 
%At the time of the experiment, the dataset comprised 2,801 unique political articles with validated negated counterparts. 
The dataset continues to self-update for future research use.
%\subsection{Further Descriptive Statisics on Amazon Reviews and Information on MediaSum News datasets}
\subsection{Descriptive Statisics on Amazon Reviews and Information on MediaSum News datasets}
\begin{table}[h]
\centering
\footnotesize
\resizebox{\linewidth}{!}{%
\begin{tabular}{lccccc}
\toprule
Dataset & Mean & Median & SD & Min & Max \\
\midrule
Amazon Reviews & 2180.3 & 2185.5 & 354.5 & 790 & 3169 \\
MediaSum News  & 1166.1 & 1054.0 & 621.8 &  89 & 2924 \\
\bottomrule
\end{tabular}}
\caption{Descriptive statistics of text lengths (tokens) for Amazon Reviews and MediaSum News.}
\label{app:length_stats}
\end{table}

\subsection{Statistical Test Results for Primacy Bias}
\label{app:stat-tests}
\begin{table}[h]
    \centering
    \scriptsize
    \begin{tabular}{llrr}
    \toprule
    \textbf{Model} & \textbf{Dataset} & \textbf{T-Statistic} & \textbf{P-Value} \\
    \midrule
    Llama-3-8B-Instruct & Amazon Reviews & 28.2896 & 8.62e-130 \\
    Llama-3-8B-Instruct & MediaSum News & 26.8965 & 2.64e-120 \\
    Llama-3.2-3B-Instruct & Amazon Reviews & 28.3100 & 6.26e-130 \\
    Llama-3.2-3B-Instruct & MediaSum News & 25.1535 & 1.50e-108 \\
    Phi-3-mini-4k-Instruct & Amazon Reviews & 18.2547 & 1.81e-64 \\
    Phi-3-mini-4k-Instruct & MediaSum News & 19.9294 & 1.14e-74 \\
    Qwen3-4B-Instruct & Amazon Reviews & 24.9284 & 4.82e-107 \\
    Qwen3-4B-Instruct & MediaSum News & 21.6017 & 3.18e-85 \\
    Gemma-3-27B-IT & Amazon Reviews & 27.7788 & 2.64e-126 \\
    Gemma-3-27B-IT & MediaSum News & 24.1413 & 1.01e-101 \\
    \bottomrule
    \end{tabular}
    \caption{Paired t-test results comparing the cosine similarity between the summary and the beginning of the source text ($s_b$) versus the middle of the source text ($s_m$). The statistics shown are for the baseline performance of each model.}
    \label{tab:t-test-results}
\end{table}
To validate that the observed primacy effect (where summary-to-beginning similarity, $s_b$, is higher than summary-to-middle similarity, $s_m$) is a systematic bias, we perform paired t-tests for each model and dataset combination. Table \ref{tab:t-test-results} presents the detailed statistical results. $p \ll .001$ across all tested models confirms that the higher similarity scores for the beginning segment are statistically significant, and the large t-statistics further indicate a substantial effect size.

\newpage
\onecolumn
\subsection{Detailed Descriptions of Mitigation Methods}
\label{app:mitigation-methods}
%Table~\ref{tab:mitigate_all} provides a  side-by-side reference for all 18 mitigation approaches evaluated in this work, including their intended bias target(s), operational definition, key hyper-parameters, and the original citation.
\renewcommand{\arraystretch}{1.3}
\begin{table*}[h!]
\scriptsize
\setlength{\tabcolsep}{3pt}
\begin{tabularx}{\textwidth}{>{\raggedright\arraybackslash}p{0.8cm}
                                 >{\raggedright\arraybackslash}p{2.7cm}
                                 >{\raggedright\arraybackslash}X
                                 >{\raggedright\arraybackslash}p{1.3cm}
                                 >{\raggedright\arraybackslash}p{2.2cm}
                                 >{\raggedright\arraybackslash}p{2.0cm}}
\toprule
\textbf{Family} & \textbf{Method} & \textbf{Operational definition} & \textbf{Target bias} & \textbf{Hyper-parameters} & \textbf{Citation}\\
\midrule
\multirow[t]{8}{*}{\textbf{Prompt}} 
 & Self-Awareness Prompt & Prepend an explicit directive that the model remain neutral, preserve sentiment, and cover all parts of the source text. & General & -- & \cite{mair2014debiasing,echterhoff2024cognitive}\\
 & Chain-of-Thought Prompting & Model generates intermediate reasoning steps for each segment before emitting a final summary, encouraging balanced coverage. & General & -- & \cite{wei2023chainofthoughtpromptingelicitsreasoning}\\
 & Cloze-Style Prompt & Insert tags \texttt{BEGIN:\_\_}, \texttt{MIDDLE:\_\_}, \texttt{END:\_\_}; model fills blanks, then emits \texttt{FINAL\_SUMMARY}. & Primacy & Tags inside prompt; no extra threshold. & \cite{liu2019hierarchicaltransformersmultidocumentsummarization}\\
 & Cognitive Counterfactual Simulation & Draft $\rightarrow$ imagine how primacy, recency, or framing bias would distort it $\rightarrow$ rewrite to avoid those distortions. & Primacy, \, Framing & Two passes. & \cite{dong2023counterfactualdebiasinggeneratingfactually}\\
 & Self-Help Debias & Draft, model self-critiques positional coverage, rewrites summary. & Primacy & Rewrite pass 300 tokens. & \cite{echterhoff2024cognitive}\\
 
 & Task-Specific CoT Prompt Calibration & Establishes clear evaluation criteria by explicitly specifying when to return ``True'' or ``False'' during fact-checking tasks. For example: ``Return true only if the described event has occurred, or if it is a direct consequence of a previously known event''.
Return false in all other cases & Hallucination & Chain-of-Thought sequence pertaining to your task & \cite{wei2023chainofthoughtpromptingelicitsreasoning}\\
 & Knowledge Boundary Awareness & Explicitly defines the model's knowledge cutoff date to create clear temporal boundaries for what the model can reasonably be expected to know. Prevents models from fabricating information about events beyond their training data by acknowledging these boundaries in prompts. & Hallucination & Knowledge cutoff date specification for the model being employed & --\\
 & Epistemic Tagging & Requires models to express confidence levels alongside factual assertions, creating a more nuanced representation of the model's knowledge state. Forces the model to evaluate its own fact-checking abilities before responding and provides users with meta-cognitive signals about response reliability. & Hallucination & Confidence level scales (we used high and low). This can include moderate, very high, very low depending on task. & \cite{lin2022teaching}\\
\midrule
\multirow[t]{3}{*}{\textbf{Chunk}} 
 % & Multi-Turn Chunking & Split into $k$ sequential chunks, summarise each, then merge. & Primacy & $k=3$; merge prompt 500 tokens. & \cite{liu2019hierarchicaltransformersmultidocumentsummarization,chang2024booookscoresystematicexplorationbooklength}\\
 & Partial Summaries Ensemble & Split the document into equal-length chunks and summarize each chunk independently (same generation hyperparameters). In a second pass, we merge the partial summaries into a final summary. & Primacy & Chunk size $\lfloor|D|/3\rfloor$; greedy merge. & \cite{ou2025contextawarehierarchicalmerginglong}\\
 & Weighted Summaries & We pre-allocate the summary length budget so that 33\% focuses on the beginning, 34\% on the middle, and 33\% on the end. The model is instructed not to exceed these chunk limits, which helps prevent overemphasis on earlier parts. & Primacy & Token ratio \(0.33{:}0.34{:}0.33\). & inspired by \cite{liu2019hierarchicaltransformersmultidocumentsummarization}\\
\midrule
\multirow[t]{2}{*}{\textbf{Re-rank}} 
 & Attention-Sort Re-ordering & Run a forward pass 
 %(without generating the final summary) 
 to estimate cross-attention weights for each paragraph or segment, then reorder segments from lowest-attended to highest-attended. The model is then re-prompted with this new order, which pushes under-attended parts later in the context and encourages more balanced coverage. & Primacy & 2 iterations; paragraph granularity. & --\\
 & Position-Invariant Shuffle & Shuffle entire sentences (split by periods) to randomize their order before prompting the model, so it cannot rely on absolute position. This can disrupt semantic continuity slightly, but ensures that coverage does not depend on text location. Diagnostic ablation for positional sensitivity. & Primacy & Period-split; seed $42$. & \cite{wang2024voomrobustvisualobject,liu2024lost}\\
\midrule
\multirow[t]{6}{*}{\textbf{Decode}} 
 & Forced Balanced Coverage & We measure coverage of the beginning, middle, and end sections via TF-IDF \cite{SALTON1988513}, and add $\log\gamma$ ($\gamma=1.5$) to logits of tokens from under-covered sections until $\lvert s_b\!-\!s_e\rvert\le0.05$. & Primacy & Threshold $0.05$; boost $\gamma=1.5$. & \cite{see2017pointsummarizationpointergeneratornetworks}\\
 & Weighted Token Decoding & Multiply next-token probabilities by weights $w_i$ (down-weight negative words, up-weight middle keywords). & Framing & $w_{\text{neg}}=0.3$, $w_{\text{mid}}=2.0$. & \cite{liu2021dexpertsdecodingtimecontrolledtext,dathathri2020plugplaylanguagemodels}\\
 & Mirostat Decoding & After each token, compute its surprise \(s_t=-\log p_t\); update the running state with \(\mu_{t+1}=\mu_t-\eta\,(s_t-\mu^\star)\), set temperature \(T_t=\exp(\mu_{t+1})\), and rescale logits as \(\mathrm{softmax}(z/T_t)\). The feedback loop keeps the observed surprise near the target \(\mu^\star\), smoothing positional coverage. & Primacy & \(\mu^\star=2.0,\;\eta=0.1\) & \cite{basu2021mirostatneuraltextdecoding}\\
 & Rejection Sampling & If top-1 token would increase chunk imbalance, set its logit to $-\infty$ and resample within top-$k$. & Primacy & $k=5$. & --\\
 & Self-Debias Decoding & Dual-pass decoding: at each step run a second forward pass on the current context preceded by a bias-inducing prefix that names the undesired attribute; obtain $p_{\text{bias}}$, compute $\Delta=p_{\text{main}}-p_{\text{bias}}$, and down-scale tokens with $\Delta<0$ via $\alpha(\Delta)=e^{\lambda\Delta}$. & Framing & $\lambda=10$; bias prefix $<\!30$ tokens; refresh every 4 steps. & \cite{schick2021selfdiagnosisselfdebiasingproposalreducing}\\
 & Local-Explanation Guard & After each token the model explains its choice; if explanation says “ignoring middle’’ or “flipping sentiment’’, reject token and resample. & Primacy, \, Framing & Check every 5 steps; rule-based detector. & --\\
\bottomrule
\end{tabularx}
\caption{Overview of the 18 bias-mitigation strategies in this work. Methods are grouped family, annotated with the bias they aim to address, and accompanied by  main hyper-parameters and primary citation(s).}
\label{tab:mitigate_all}
\end{table*}
\clearpage

\subsection{Full Mitigation Results}
\label{sec:extended_results}
%Table~\ref{tab:combined} reports the complete results for all
%models and mitigation strategies.
\begin{table*}[h!]
\scriptsize
\centering
\setlength{\tabcolsep}{2.7pt}
\begin{tabular}{lccccccccccccc}
\toprule
& \multicolumn{6}{c}{\textbf{Amazon Reviews}} & \multicolumn{6}{c}{\textbf{MediaSum News}} \\
\cmidrule(lr){2-7}\cmidrule(lr){8-13}
\textbf{Method} &
\(\fphi\downarrow\) &
\(\bar{s}_b \uparrow\) & \(\bar{s}_m \uparrow\) & \(\bar{s}_e \uparrow\) &
$\fpsi\downarrow$ &
\(\rho_{\text{pri}}\downarrow\) &
\(\fphi\downarrow\) &
\(\bar{s}_b \uparrow\) & \(\bar{s}_m \uparrow\) & \(\bar{s}_e \uparrow\) &
$\fpsi\downarrow$ &
\(\rho_{\text{pri}}\downarrow\) \\
\midrule
%\multicolumn{13}{c}{\textbf{GPT-3.5-turbo}}\\
\midrule
\textbf{GPT-3.5-turbo} & 16.0\% & 0.848 & 0.826 & 0.825 & 7.6\% & -- & 24.8\% & 0.840 & 0.823 & 0.820 & 6.1\% & -- \\
\midrule
%\multicolumn{13}{c}{\textbf{Llama-3.2-3B-Instruct}}\\
\midrule
\textbf{Llama-3.2-3B-Instruct} & \textbf{14.9\%} & 0.860 & 0.842 & 0.840 & \textbf{7.4\%} & -- & 22.1\% & 0.851 & 0.837 & 0.832 & \textbf{5.1\%} & -- \\
\midrule
% Multi-Turn Chunking & 22.0\% & 0.889 & 0.874 & 0.874 & 13.8\% & 67.3\% & 28.4\% & 0.876 & 0.864 & 0.857 & 10.8\% & 68.5\% \\
Self-Awareness Prompt & 24.0\% & 0.908 & 0.883 & 0.881 & 19.7\% & 80.4\% & \textbf{20.9\%} & 0.907 & 0.883 & 0.875 & 15.7\% & 85.0\% \\
Chain-of-Thought & 23.6\% & 0.909 & 0.883 & 0.881 & 19.9\% & 81.1\% & 22.6\% & 0.907 & 0.883 & 0.876 & 16.9\% & 85.9\% \\
Cloze-Style Prompt & 23.8\% & 0.908 & 0.883 & 0.881 & 19.8\% & 81.0\% & 22.6\% & 0.907 & 0.883 & 0.876 & 16.9\% & 85.9\% \\
Cognitive Counterfactual Sim. & 23.7\% & 0.908 & 0.883 & 0.881 & 20.2\% & 80.8\% & 22.5\% & 0.907 & 0.884 & 0.876 & 16.6\% & 84.8\% \\
Self-Help Debias & 23.0\% & 0.908 & 0.883 & 0.881 & 20.6\% & 81.3\% & 23.6\% & \textbf{0.908} & \textbf{0.884} & \textbf{0.876} & 15.9\% & 85.8\% \\
Partial Summaries Ensemble & 21.0\% & 0.886 & 0.870 & 0.869 & 14.8\% & 69.1\% & 27.6\% & 0.871 & 0.854 & 0.846 & 14.3\% & 78.2\% \\
Weighted Summaries & 20.7\% & \textbf{0.921} & \textbf{0.903} & 0.899 & 15.4\% & 75.6\% & 29.4\% & 0.876 & 0.862 & 0.852 & 9.8\% & 80.1\% \\
Attention-Sort Re-ordering & 19.8\% & 0.905 & 0.886 & 0.885 & 16.8\% & 71.4\% & 26.4\% & 0.859 & 0.841 & 0.830 & 14.6\% & 78.5\% \\
Position-Invariant Shuffle & 26.6\% & 0.903 & 0.890 & 0.889 & 10.8\% & \textbf{67.1\%} & 30.8\% & 0.875 & 0.868 & 0.859 & 8.9\% & \textbf{68.4\%} \\
Forced Balanced Coverage & 21.0\% & 0.905 & 0.884 & 0.883 & 17.5\% & 72.5\% & 26.2\% & 0.891 & 0.867 & 0.859 & 19.0\% & 83.0\% \\
Weighted Token Decoding & 18.9\% & 0.865 & 0.846 & 0.844 & 17.3\% & 71.0\% & 29.6\% & 0.855 & 0.841 & 0.834 & 14.4\% & 72.3\% \\
Mirostat Decoding & 20.0\% & 0.918 & 0.903 & \textbf{0.901} & 12.2\% & 70.0\% & 31.2\% & 0.901 & 0.882 & 0.875 & 11.7\% & 79.1\% \\
Rejection Sampling & 19.8\% & 0.904 & 0.885 & 0.883 & 16.7\% & 72.0\% & 23.8\% & 0.890 & 0.865 & 0.857 & 19.5\% & 82.7\% \\
Self-Debias Decoding & 22.9\% & 0.895 & 0.879 & 0.878 & 14.3\% & 68.9\% & 26.0\% & 0.879 & 0.854 & 0.847 & 17.6\% & 83.9\% \\
Local-Explanation Guard & 20.2\% & 0.865 & 0.844 & 0.842 & 17.9\% & 73.5\% & 26.5\% & 0.870 & 0.847 & 0.836 & 20.1\% & 81.8\% \\
\midrule
%\multicolumn{13}{c}{\textbf{Llama-3-8B-Instruct}}\\
\midrule
\textbf{Llama-3-8B-Instruct} & 14.5\% & 0.855 & 0.837 & 0.836 & \textbf{7.0\%} & -- & 21.9\% & 0.847 & 0.834 & 0.828 & \textbf{4.0\%} & -- \\
\midrule
% Multi-Turn Chunking & 18.1\% & 0.907 & 0.888 & 0.886 & 17.4\% & 72.0\% & 28.6\% & 0.883 & 0.868 & 0.856 & 13.4\% & 76.4\% \\
Self-Awareness Prompt & 15.2\% & 0.909 & 0.881 & 0.878 & 22.9\% & 83.2\% & \textbf{21.0\%} & 0.889 & 0.864 & 0.855 & 19.7\% & 85.5\% \\
Chain-of-Thought & 15.3\% & 0.909 & 0.881 & 0.878 & 22.6\% & 82.7\% & 22.5\% & 0.890 & 0.865 & 0.856 & 19.2\% & 85.6\% \\
Cloze-Style Prompt & 15.2\% & 0.909 & 0.881 & 0.878 & 21.9\% & 82.4\% & 21.9\% & 0.891 & 0.866 & 0.856 & 20.2\% & 85.3\% \\
Cognitive Counterfactual Sim. & 14.7\% & 0.909 & 0.881 & 0.878 & 22.5\% & 82.7\% & 21.8\% & 0.890 & 0.866 & 0.856 & 19.0\% & 84.8\% \\
Self-Help Debias & 15.2\% & 0.909 & 0.881 & 0.879 & 21.8\% & 83.0\% & 21.7\% & 0.891 & 0.865 & 0.856 & 19.2\% & 85.4\% \\
Partial Summaries Ensemble & 16.5\% & 0.890 & 0.871 & 0.870 & 15.4\% & 72.7\% & 28.5\% & 0.874 & 0.858 & 0.849 & 14.1\% & 79.3\% \\
Weighted Summaries & 15.1\% & \textbf{0.925} & \textbf{0.904} & \textbf{0.902} & 15.1\% & 77.0\% & 26.6\% & 0.880 & 0.866 & 0.855 & 9.0\% & 81.8\% \\
Attention-Sort Re-ordering & 16.9\% & 0.879 & 0.858 & 0.857 & 17.5\% & 74.7\% & 25.0\% & 0.869 & 0.845 & 0.831 & 20.7\% & 85.0\% \\
Position-Invariant Shuffle & 21.2\% & 0.899 & 0.888 & 0.887 & 11.5\% & \textbf{64.4\%} & 27.0\% & 0.861 & 0.854 & 0.847 & 8.0\% & \textbf{66.3\%} \\
Forced Balanced Coverage & 15.6\% & 0.900 & 0.878 & 0.877 & 19.5\% & 74.2\% & 23.8\% & 0.862 & 0.839 & 0.832 & 18.1\% & 81.3\% \\
Weighted Token Decoding & \textbf{13.6\%} & 0.871 & 0.851 & 0.852 & 18.4\% & 70.0\% & 24.9\% & 0.852 & 0.831 & 0.823 & 18.1\% & 77.3\% \\
Mirostat Decoding & 13.8\% & 0.916 & 0.896 & 0.895 & 15.5\% & 75.8\% & 31.3\% & \textbf{0.900} & \textbf{0.875} & \textbf{0.867} & 18.3\% & 83.0\% \\
Rejection Sampling & 14.3\% & 0.911 & 0.890 & 0.889 & 18.9\% & 73.9\% & 26.8\% & 0.872 & 0.849 & 0.842 & 18.5\% & 80.0\% \\
Self-Debias Decoding & 16.0\% & 0.905 & 0.883 & 0.882 & 18.8\% & 75.1\% & 25.5\% & 0.879 & 0.855 & 0.848 & 18.2\% & 82.8\% \\
Local-Explanation Guard & 16.1\% & 0.859 & 0.837 & 0.836 & 18.9\% & 74.6\% & 25.9\% & 0.852 & 0.830 & 0.822 & 18.3\% & 79.1\% \\
\midrule
%\multicolumn{13}{c}{\textbf{Phi-3-mini-4k-Instruct}}\\
\midrule
\textbf{Phi-3-mini-4k-Instruct} & 34.5\% & 0.836 & 0.822 & 0.823 & 5.4\% & -- & 26.2\% & 0.828 & 0.827 & 0.823 & \textbf{4.9\%} & -- \\
\midrule
% Multi-Turn Chunking & 28.2\% & 0.890 & 0.876 & 0.872 & 13.9\% & 69.0\% & 30.6\% & 0.855 & 0.845 & 0.839 & 12.1\% & 67.8\% \\
Self-Awareness Prompt & 25.5\% & 0.882 & 0.866 & 0.867 & 14.5\% & 69.5\% & \textbf{24.9\%} & 0.850 & 0.832 & 0.826 & 16.6\% & 73.7\% \\
Chain-of-Thought & 23.9\% & 0.884 & 0.868 & 0.868 & 14.1\% & 69.7\% & 26.0\% & 0.851 & 0.832 & 0.827 & 15.7\% & 73.5\% \\
Cloze-Style Prompt & 24.7\% & 0.883 & 0.868 & 0.868 & 13.6\% & 67.7\% & 25.9\% & 0.850 & 0.832 & 0.826 & 16.4\% & 73.5\% \\
Cognitive Counterfactual Sim. & 22.8\% & 0.877 & 0.862 & 0.862 & 14.4\% & 67.7\% & 25.5\% & 0.850 & 0.832 & 0.826 & 16.2\% & 73.5\% \\
Self-Help Debias & 22.8\% & 0.879 & 0.863 & 0.864 & 14.3\% & 68.5\% & 25.1\% & 0.851 & 0.832 & 0.826 & 16.0\% & 75.5\% \\
Partial Summaries Ensemble & 28.0\% & 0.893 & 0.880 & 0.875 & 11.5\% & 71.3\% & 27.7\% & 0.858 & 0.845 & 0.836 & 11.4\% & 73.9\% \\
Weighted Summaries & \textbf{22.6\%} & \textbf{0.920} & \textbf{0.904} & \textbf{0.899} & 11.2\% & 78.1\% & 27.8\% & \textbf{0.880} & \textbf{0.871} & \textbf{0.863} & 7.4\% & 72.1\% \\
Attention-Sort Re-ordering & 33.1\% & 0.887 & 0.877 & 0.876 & 9.9\% & 66.8\% & 28.1\% & 0.837 & 0.823 & 0.818 & 14.5\% & 67.9\% \\
Position-Invariant Shuffle & 28.2\% & 0.877 & 0.866 & 0.865 & 9.3\% & 67.6\% & 28.2\% & 0.837 & 0.834 & 0.825 & 7.7\% & 63.0\% \\
Forced Balanced Coverage & 31.5\% & 0.901 & 0.891 & 0.893 & 7.6\% & 61.5\% & 29.9\% & 0.844 & 0.831 & 0.830 & 11.8\% & 66.3\% \\
Weighted Token Decoding & 25.0\% & 0.860 & 0.849 & 0.854 & 9.8\% & 56.9\% & 28.9\% & 0.841 & 0.829 & 0.828 & 14.4\% & 63.0\% \\
Mirostat Decoding & 33.2\% & 0.891 & 0.883 & 0.884 & \textbf{4.2\%} & 58.8\% & 34.3\% & 0.854 & 0.845 & 0.843 & 8.7\% & 61.3\% \\
Rejection Sampling & 22.7\% & 0.885 & 0.876 & 0.878 & 8.2\% & 60.5\% & 28.3\% & 0.853 & 0.836 & 0.835 & 14.6\% & 66.8\% \\
Self-Debias Decoding & 28.8\% & 0.894 & 0.885 & 0.888 & 7.1\% & \textbf{56.4\%} & 31.3\% & 0.844 & 0.832 & 0.834 & 11.3\% & \textbf{60.0\%} \\
Local-Explanation Guard & \textbf{22.5\%} & 0.886 & 0.875 & 0.879 & 9.4\% & 59.7\% & 27.3\% & 0.841 & 0.828 & 0.826 & 12.4\% & 65.8\% \\
\bottomrule
\end{tabular}
\caption{
\small \textbf{Coverage, framing-change, and primacy-bias metrics} for Amazon Reviews and MediaSum on $\fphi$ (framing-change fraction, $\downarrow$), 
$\bar{s}_b,\bar{s}_m,\bar{s}_e$ (cosine similarity with the first/middle/last third of the source, $\uparrow$), 
$\fpsi$ (share of summaries whose similarity to the beginning exceeds that to the middle by $>5\%$, $\downarrow$), 
and $\rho_{\text{pri}}$ (share of summaries exhibiting primacy bias, $\downarrow$).  
% For clarity the table is \emph{filtered}: for each of the four base LLMs we keep only the baseline plus the few mitigation methods that achieve the best score in at least one of these columns.
}
\label{tab:combined}
\end{table*}

\clearpage
\newpage
\subsection{Prompt Templates}\label{app:prompt-strategies}
%Table~\ref{tab:prompt_strategies} lists the exact prompt instruction templates supplied to each
%model under all the experimental settings.
\begin{table*}[h!]
\scriptsize
\centering
\setlength{\tabcolsep}{3pt}
\begin{tabular}{p{3.5cm} p{10.5cm}}
\toprule
\textbf{Prompt Strategy} & \textbf{Prompt} \\
\midrule
\multicolumn{2}{c}{\textbf{Summarization and General Bias Mitigation Prompts}} \\
\midrule
Self-Awareness Prompting &
``You are an unbiased summarizer. Be mindful not to introduce any framing bias or omit the middle. Preserve the original sentiment. %\newline 
Please summarize the following text: \texttt{[DOCUMENT\_TEXT]} \newline FINAL\_SUMMARY:'' \\
\addlinespace[0.4em]
Chain-of-Thought (Summarization) &
``Please read the text below carefully. Then break down the text into beginning, middle, and end, describing each portion in detail. After that, produce a final summary. Use the following format: \newline BEGIN\_ANALYSIS: [describe the beginning] \newline MIDDLE\_ANALYSIS: [describe the middle] \newline END\_ANALYSIS: [describe the end] \newline FINAL\_SUMMARY: [your final concise summary] \newline Text: \texttt{[DOCUMENT\_TEXT]}'' \\
\addlinespace[0.4em]
Cloze-Style Prompt &
``Fill the blanks for each part of the text: \newline BEGIN: \_\_\_\_  %\newline
MIDDLE: \_\_\_\_ %\newline 
END: \_\_\_\_ \newline Text: \texttt{[DOCUMENT\_TEXT]} \newline When you fill them in, finally produce: \newline FINAL\_SUMMARY:'' \\
\addlinespace[0.4em]
Cognitive Counterfactual Simulation (Final Revision) &
``Original text: \texttt{[DOCUMENT\_TEXT]} \newline Initial Draft: \texttt{[DRAFT\_SUMMARY]} \newline Here are the potential biased deviations: \texttt{[LIST\_OF\_SIMULATED\_BIAS\_DEVIATIONS]} \newline Please produce a final summary that avoids these biases. \newline FINAL\_SUMMARY:'' \\
\addlinespace[0.4em]
Self-Help Debias (Rewrite Stage) &
``Rewrite the following summary to remove any bias, preserve the original sentiment, and ensure equal coverage of beginning, middle, and end: \texttt{[DRAFT\_SUMMARY]} \newline FINAL\_SUMMARY:'' \\
\addlinespace[0.4em]
Partial Summaries Ensemble (Merge Prompt) &
``Combine the following partial summaries into one coherent summary: \newline \texttt{[PARTIAL\_SUMMARY\_1]} %\newline 
\texttt{[PARTIAL\_SUMMARY\_2]} %\newline 
... \newline FINAL\_SUMMARY:'' \\
\addlinespace[0.4em]
Weighted Summaries (Chunk Prompt) &
``Summarize this portion in about \texttt{[PORTION\_TOKEN\_BUDGET]} tokens: \newline \texttt{[CHUNK\_TEXT]} \newline FINAL\_SUMMARY:'' \\
\addlinespace[0.4em]
Attention-Sort Re-ordering (Final Prompt) &
``CONTEXT: \newline Segment 1: \texttt{[SORTED\_SEGMENT\_1\_TEXT]} \newline Segment 2: \texttt{[SORTED\_SEGMENT\_2\_TEXT]} \newline ... \newline TASK: Summarize: \newline FINAL\_SUMMARY:'' \\
\addlinespace[0.4em]
Position-Invariant Shuffle (Prompt) &
``The text is out of order; please summarize it fully: \texttt{[SHUFFLED\_DOCUMENT\_TEXT]} \newline FINAL\_SUMMARY:'' \\
\midrule
\multicolumn{2}{c}{\textbf{Hallucination Mitigation Prompts}} \\
\midrule
Baseline &
``You are a fact-checking AI. For each statement, predict if the event has occured or not and respond with ONLY 'True' if it's factual or 'False' if it's not. Do not provide any explanation. Respond with TRUE or FALSE only and nothing else.'' \\
\addlinespace[0.4em]
Prompt Calibration (CoT) &
``You are a fact-checking AI. Follow these strict guidelines: \newline
- Respond with 'True' ONLY if: \newline
\phantom{--} * You are 100\% certain the statement describes an event that has occurred \newline
\phantom{--} * You have clear knowledge of the event from your training data \newline
\phantom{--} * You can identify specific details confirming the event happened \newline
- Respond with 'False' if: \newline
\phantom{--} * You have ANY doubt about the statement's accuracy \newline
\phantom{--} * The statement contains ANY misleading or incorrect information \newline
\phantom{--} * You lack reliable information about the statement \newline
Respond with EXACTLY one word: 'True' or 'False'. No explanations.'' \\
\addlinespace[0.4em]
Knowledge Boundary Awareness &
``You are a fact-checking AI with knowledge up to \texttt{\{knowledge\_cutoff\}}. For each statement, predict if the event has occurred or not. We check your decision-making abilities on futuristic events based on current data. Respond with only 'True' or 'False' for each news event. \newline If the event is scheduled to occur after \texttt{\{knowledge\_cutoff\}}, make a good prediction based on your knowledge up to \texttt{\{knowledge\_cutoff\}}. \newline RESPOND WITH EXACTLY ONE WORD: 'True' or 'False'.'' \\
\addlinespace[0.4em]
Epistemic Tagging &
``You are a fact-checking AI. Analyze this statement and respond with: \newline
- 'True [High Confidence]' or 'True [Low Confidence]' \newline
- 'False [High Confidence]' or 'False [Low Confidence]' \newline
Use these epistemic tags to guide your reasoning: \newline
[Certain] = Information you know with high confidence based on your training data \newline
[Uncertain] = Information you're less confident about or may be outside your knowledge \newline
[Reasoning] = Step by step analysis to determine veracity \newline
[Conclusion] = Your final determination with confidence level \newline
Respond ONLY with one of the four exactly formatted options listed above and nothing else.'' \\
\bottomrule
\end{tabular}
\caption{Prompt strategies and the prompts used in our experiments.}
\label{tab:prompt_strategies}
\end{table*}

\pagebreak
\subsection{Experimental Details on Products for Human Study}
\scriptsize
\begin{longtable}
{p{3cm}p{7cm}p{5cm}}
\label{Table::Reviews} \\
\toprule
Product & Review & Summary \\
\midrule
\endfirsthead

\multicolumn{3}{c}%
{{\bfseries Table \thetable\ continued}} \\
\toprule
Product & Review & Summary \\
\midrule
\endhead

\midrule
\multicolumn{3}{r}{{Continued on next page}} \\
\endfoot

\bottomrule
\endlastfoot
%Product & Review & Summary\\
%\midrule
Refrigerator water filter &
We installed this filter on our Whirlpool Refrigerator and two months later the water was only trickling. I reseated it just in case something had gotten in the line. Nothing had gotten in. The filter was plugged after two months. The valve and all are fine. After reseating water was still trickling. We ordered and installed the factory Whirlpool filter and as expected the water flow is normal. The waster valve made noise in the days that this cheap filter was clogged, think of the damage that does over time! Why take a chance with this cheaper unit coming apart inside and risking a Service Call because it damaged the valve that is hard to get to?????<br /><br />PAY TYHE MONEY. YOU BOUGHT A NICE FRIDGE. BUY THE WHIRLPOOL FACTORY FILTER.  IT IS YOUR HEALTH, IF THIS FILTER CLOGGS AFTER TWO MONTHS THEN IT IS FAILING INSIDE. The old whirlpool filter went over one year not changed and never had an issue. I now have the Whirlpool filter in my subscribe-n-save every six months.
&
I initially tried a cheaper filter for my Whirlpool Refrigerator, but after two months it clogged and caused water flow issues. I switched to the factory Whirlpool filter and the water flow returned to normal. It's worth investing in the quality filter to avoid potential damage and service calls.
\\
\midrule
Coffee water filter &
Three stars ONLY because the filter assembly was almost impossible for me to get open. If you can't replace the filter, the filter assembly is useless. The bottom of the filter assembly requires the sides of the hard acrylic be squeezed together in order for the bottom to be released (refer to the attached photo). After several failed attempts, I was able to get the bottom released by placing it against a hard object and pressing down on the opposite side with both thumbs. Very worrisome because the acrylic could just as easily cracked, ruining the filter assembly. So, the bottom line is if you don't have strong hands, it will be very hard to open the filter assembly to replace the filter every two months. Otherwise, this starter kit fits nicely in my Keurig 2.0, K300,  the water is clean and tastes very good. The reusable K-cup is another plus for the starter kit. My Keurig recognizes it and allows for brewing loose leaf tea or non-K-cup coffee of my choice.
&
The filter assembly was difficult to open, requiring strong hands to avoid cracking the acrylic. However, the water tastes great and the reusable K-cup is a nice addition for brewing loose leaf tea or non-K-cup coffee. Fits well in my Keurig 2.0, K300.
\\
\midrule
Electric kettle &
Cons:<br />  * The first thing you will notice on unpacking is the short cord. It is the shortest appliance cord I have ever seen and is quite inconvenient.<br />  * The build of the kettle is not great, especially considering the price. I am worried that the gooseneck will fall off.<br />  * The electronic controls are clumsy when compared to the Cuisinart Perfect Temp simple presets.<br />  * The heating element is small so heating is slower than expected.<br />  * Bonavita designs never get it quite right. I use their coffeemaker which works great but has a smoked glass water reservoir that makes it hard to get an exact water level. The waterproof cover for the electronics on this kettle are frosted instead of clear so they are not easy to see.<br /><br />Pros:<br />  * Heats water to very close to the desired temps.<br />  * Gooseneck is great for pourover coffee preparation.<br />  * Takes up little counter space but must be very close to an outlet.
&
The short cord and build quality are disappointing, but the Bonavita kettle heats water well and is compact. The gooseneck is perfect for pour-over coffee. Controls are not as user-friendly as other brands. Overall, it's a decent option for those who prioritize temperature accuracy and pour control.
\\
\midrule
Screen protector &
Not overwhelmed nor am I disappointed. This protector is on par with many on the market. Installation is all about preparation. The included guide frame is nice and takes guesswork out of alignment. If you’ve prepped the screen and manage to not stir up dust then you should get an installation that only needs bubbles squeezed out.  Take your time and you can get a perfect installation.  Take your time.<br /><br />That being said, the frame is all that really sets this one apart from others for this iPad and the others we own. You will get the same quality as if you pick another brand as they’re all pretty much the same one in different packaging.<br /><br />It works and looks as it should. After a bit of use you’ll see fingerprints start to show up. It will happen. You can minimize this by using a clean soft cloth with no cleaner. Eventually though it will start to show every print. Up to you to live with it or install the second one.
&
This screen protector is decent, with easy installation using the guide frame. Quality is similar to other brands. Fingerprints show up after use, but can be minimized with a clean cloth. Overall, it works as expected but may require a second installation for a flawless look.\\
\midrule
Dryer vent hood &
The Good:<br />The construction is excellent.  Thick stainless steel and well made.<br />The damper works well. We are using this as a dryer vent and the damper moves out fully with even the mildest air flow. Its great not to have a screen that clogs with lint since this is installed on the second floor and getting to the old vent was a big hassle.<br /><br />The downsides:<br />1. The "flapper" oval foam seal that reduces noise and, more importantly, seals the vent when closed came off immediately.  With the first use, it went flying off. The important thing is that without this foam, there is a decent sized gap at the top of the flapper that bugs and outside air could get into. I reattached it with some silicon and hopefully that does the trick.  The manufacturer needs to use a better adhesive.<br />2. The rubber seal at the base makes installation a challenge.  It is installed with friction and comes off with the slightest bump. I had to make quite an effort to keep it in place during install. Note: I wish I had glued it on the day before, so it stayed in place during install rather than fighting it.
&
The stainless steel construction is top-notch and the damper works perfectly for dryer vent use. However, the foam seal came off easily and the rubber seal at the base made installation difficult. Overall, a great product with minor flaws that can be easily fixed with some DIY solutions.\\
\midrule
Video doorbell &
I've had a lot of frustration with this product and was considering returning it. It required a lot of troubleshooting after installation based on what was within its view. I suspect the original firmware had problems with audio/video quality that was fixed with a factory reset and registration as a new doorbell. A WiFi router sets about five feet behind the doorbell. Motion detection was triggered at sunrise every day, waking my dog and I up, and I was faced with a puppy who wanted to interact with me an hour before my alarm went off. This was fixed a few weeks later by installing a wedge kit to angle the camera so a wall was not in view.<br /><br />Pros - Easy setup. Get notified by phone. Easy to download video of events. Able to talk with people. Can setup active motion zones. Night vision. Fish eye lens covers a large area. Can snooze motion detection while you cut grass or hang out on the porch. Can set schedule of when device operates. Friendly staff via email/twitter.<br /><br />Cons - Initial problems with audio/video quality. Walls trigger motion detection at sunrise regardless of motion zones setup within the app. Limited choice of sounds on device and smartphones.
&
I initially had issues with this product but after troubleshooting and adjusting settings, it improved. Easy setup, notifications, video download, two-way talk, motion zones, night vision, wide coverage, snooze feature, scheduling, and helpful customer service. However, initial audio/video problems and motion detection triggered by walls at sunrise were frustrating. Limited sound options.\\
\midrule
Radio &
Overall it’s a well working radio. Sound is decent and clear. Display is nice looking. However the only knob is a multi function nightmare. Gone are the days of just reaching over to adjust squelch or volume with a quick turn of a knob. You got to choose the function then turn knob. Really irritating at least. The function knob is also very stiff and doesn't turn smoothly. This radio is getting pulled out of main driver and thrown into my pickup. I really don’t like it. A shame Anytone was unable to give you two knobs, Volume and Squelch. Recommendation: Spend an extra hundred and buy a better radio.<br />Follow-up: 9/2/18 - I decided to give it a second chance. I pulled this radio off the shelf and immersed myself into programming some frequencies by hand. Took some time figuring out the quirks, like you need to program UHF frequencies when your on B radio. If you try to do it on A, your split will be 600 mhz instead of 5 khz. Well after all this learning I decided to mount it in my car. Its really not to bad once its programmed. I would however recommend a larger external speaker though since the small size radio just cant put out good sound. The other irritating thing is Channel names are limited to 5 characters. Yikes.
&
Initially disappointed with the single function knob, but after giving it a second chance and learning its quirks, I found it to be a decent radio once programmed. Sound quality is lacking, but overall not bad for the price. Would recommend a larger external speaker for better sound.\\
\midrule
Reusable K-Cup &
This isn't compatible with some machines, out of the box. Keurig has been increasing the size of the top needle, or possibly moving it slightly off-center on some models. When using a K-Cup, that never matters of course. When using a reusable K-Cup like this one, the machine's lid might not close without ruining the machine ... or without wearing down either the needle or the reusable K-Cup's lid opening.<br /><br />They're all copying each other's designs, because of course they are. Who's gonna be brave, and the first to just make the top opening a radical 1 mm wider?<br /><br />The "smoosh down" method of securing the lid isn't the best, however I'm aware they're all like that. If using a disposable filter, push it down into the cup, otherwise the lid will not stay shut. Accept maybe a little bit of sediment in your coffee mug as a result. Yeah, I know, that's what a disposable paper filter is supposed to mitigate.
&
This reusable K-Cup may not fit all machines perfectly, but works fine with K-Cups. The design is similar to others, but could benefit from a wider top opening. The lid may need extra pressure to close, especially with a disposable filter. Overall, it gets the job done with minor inconveniences.\\
\midrule
Headlamp &
I got this headlamp mainly for trail running in the dark. It's a bright light that has more of a surround light. There are different settings for brightness. There is also the ability to turn on just the center lights which is better for focusing on what is in front of me on the trail. These are the pros on the light.<br /><br />Unfortunately, there are cons that are big enough that made me drop this light a couple of stars. First of all, it's stiff on my head and not really comfortable for trail running or hiking. The battery pack sits kind of heavy on the back of my head. I'm not a fan at all of the charging system with the prongs that need to line up just right. It would have been so much easier to make this a regular USB charging system that would have been compatible with other charging cords.<br /><br />This will be going in my camping trailer and used for camping. On the plus side, this will be one of the brightest headlamps I've used for camping and will make a great addition to my  camp supplies. This would also work really well for those working outside in the dark when brightness is more important than comfort.
&
This headlamp is bright with different settings, great for trail running. However, it's uncomfortable and has a finicky charging system. Despite the cons, it's perfect for camping and outdoor work where brightness is key.\\
\midrule
Headset &
I have the Jabra 35 headset also and this one is noticably different with regard to the quality of phone calls, but that's really the only material difference I can find in my day to day use. The other unit does offer some noise cancellation effects that this one doesn't have and people you call can definitely tell the difference. My husband said that he could hear background noise much more clearly when I called him from this 15 than with the 35.<br /><br />As far as battery life, this thing is really solid. I've used for three hours on two different days without recharging between days. I primarily use it to listen to audiobooks while I'm doing my barn chores and checking fence lines. I have noticed that I'm pretty limited with the distance I can stray from my physical phone before this headset begins to stutter to lose signal - it seems to be about 20 feet with my Sony Xperia phone.
&
The Jabra 15 headset has great call quality but lacks noise cancellation. Battery life is impressive, lasting three hours over two days. However, signal range is limited to about 20 feet from my phone. Overall, a solid choice for calls and audiobooks, but lacking in some features compared to the 35.\\
\bottomrule

\caption{Product reviews and their summaries used for the human-subject experiment.}
\label{Table:Reviews}
\end{longtable}

\end{document}